\documentclass[review]{elsarticle}

\usepackage{lineno,hyperref,gensymb,amsmath,subcaption,dingbat,multirow,amssymb}
\modulolinenumbers[5]
\usepackage[table,xcdraw,dvipsnames]{xcolor}
    
\definecolor{customgray}{gray}{0.8}
\journal{arxiv.org}

\begin{document}

\begin{frontmatter}

\title{On estimating gaze by self-attention \\ augmented convolutions}

\author[addressufba]{Gabriel Lefundes}
\ead{gabriel.lefundes@ufba.br}
\author[addressufba]{Luciano Oliveira\corref{cor1}}
\ead{lrebouca@ufba.br}
\cortext[cor1]{Corresponding author: Luciano Oliveira.}
\address[addressufba]{Intelligent Vision Research Lab, \\ Federal University of Bahia, Bahia, Brazil}

\begin{abstract}
Estimation of 3D gaze is highly relevant to multiple fields, including but not limited to interactive systems, specialized human-computer interfaces, and behavioral research. Although recently deep learning methods have boosted the accuracy of appearance-based gaze estimation, there is still room for improvement in the network architectures for this particular task. Therefore we propose here a novel network architecture grounded on self-attention augmented convolutions to improve the quality of the learned features during the training of a shallower residual network. The rationale is that self-attention mechanism can help outperform deeper architectures by learning dependencies between distant regions in full-face images. This mechanism can also create better and more spatially-aware feature representations derived from the face and eye images before gaze regression. We dubbed our framework ARes-gaze, which explores our \textbf{A}ttention-augmented \textbf{Res}Net (\textbf{ARes-14}) as twin convolutional backbones. In our experiments, results showed a decrease of the average angular error by 2.38\% when compared to state-of-the-art methods on the MPIIFaceGaze data set, while achieving a second-place on the EyeDiap data set. It is noteworthy that our proposed framework was the only one to reach high accuracy simultaneously on both data sets.
\end{abstract}


\end{frontmatter}

\section{Introduction}

Gaze estimation is an active area of research within computer vision, and its relevance spans a large array of fields. For instance, Gaze can be a valuable source of information in behavioral and health research \cite{behaviour-lookwhere,behaviour-walking,behaviour-autism}, augmented and virtual reality \cite{AVR-handsfree,AVR-mobilegaming,AVR-system}, mobile applications \cite{mobile-gazeinteraction, AVR-mobilegaming}, human-computer interaction \cite{gaze-gesture}, and even natural language processing (NLP) pipelines \cite{NLP-multimodal}.

Methods of gaze estimation can be categorized as model-based or appearance-based \cite{survey}. The former relies on explicitly modeling the subject's eye and using some type of feedback (usually with the need for specialized hardware such as active infrared LEDs \cite{infrared} or wearable devices \cite{wearableglasses}) to infer the gaze direction geometrically. This approach can reach accurate results in controlled environments, although suffering from hardware cost and installation overhead. Also, model-based gaze estimation is usually limited by external factors such as lighting conditions and lower tolerance for subject pose and distance. In contrast, appearance-based approaches attempt to directly predict the gaze vector from RGB images of the subject by mapping a regression function that can be ultimately learned from data.

While challenges like lighting conditions and unconstrained subject pose remain, the use of deep learning and in-the-wild large-scale data sets \cite{mpiigaze,rtgene,multiview} have greatly improved the accuracy of appearance-based methods, which in general only need simple monocular cameras as input sensors. Recent publications in the field have focused on exploring different neural network architectures and training conditions to raise the performance of the current state-of-the-art. Notably, many works have remarked that by using full-face images along with the usually extracted eye-patches as the input, can improve the prediction accuracy significantly \cite{writtenallover,itracker,dilatednet}. This is so since full-face images carry relevant information about the subject's pose. 

Here, we explore the recent trend of attention mechanisms in deep learning \cite{attentionisall} as a way to produce higher quality features by improving the spatial awareness of the network. The rationale is to better leverage the relationship between coarse pose information from face images and fine information from eye-patches.

\subsection{Related works}

\noindent\textbf{Appearance-based gaze estimation:} Early works in appearance-based gaze estimation used well-established machine-learning algorithms like adaptive linear regression \cite{adaptivelinearreg}, support vector regression \cite{svrmanifold}, and random forests \cite{synthrandomforest} to learn the mapping function from eye images to gaze vectors. Recently, convolutional neural networks (CNNs) have shown great success in gaze estimation, with its first published iteration \cite{mpiigaze} reporting significant gains over the previous state-of-the-art works. Subsequent publications have then built upon the notion of using CNNs by proposing different input models for the convolutional networks like images of the entire face and a binary grid to encode head size and position \cite{itracker}. Another explored avenue was to use different strategies to combine features extracted from the eyes and the face. Geometry constraints are used in \cite{geometryconst} to connect head pose and eye movement in an informed manner bound by physical limits of pose and movement. In \cite{coarse2fineatt}, an attention gating strategy is proposed as a way to refine gaze predictions made on full-face images adaptively by using residuals from isolated-eye images in a coarse-to-fine manner. 

Other works have proposed taking into account domain knowledge and peculiarities of the gaze estimation task while designing the architecture of the CNN itself. For example, in \cite{assymetry}, the asymmetrical nature of left and right eyes is posited to have relevance on the result of gaze estimation, and accuracy gains are reported when encoding and leveraging that asymmetry in a deep neural network. Another example of domain-specific modeling is found in \cite{dilatednet} where dilated convolutions are used as a replacement for max-pooling layers to better capture small differences in eye images. Since finer eye movement is highly relevant to gaze estimation, it can easily be lost in the downsampling stages of neural networks. In \cite{rnngaze}, recurrent CNNs are used and shown to improve prediction accuracy significantly on continuous inference. This is done so because it is plausible to consider gaze an inherently temporal phenomenon, which is grounded by the notion that where people are looking at, in a particular moment in time, directly depends on where they were looking at, in a previous moment. In \cite{writtenallover}, a spatial-weights mechanism is proposed to learn spatial importance maps and predict gaze directions using only face images as input. This map serves as a guide to the following layers of the CNN, learning to locate important features on the normalized input image (eyes, nose) while pointing to where the network focus should be. The rationale behind this approach is similar to ours, except that we estimate what would be the spatial importance in the form of multiple attention layers maps, doing so in an implicit way. This serves not only as a way to learn possible locations for important facial regions, but the use of self-attention also allows the maps to correlate these often distant features in a more abstract way not feasible for regular convolutional layers.
\newline
\newline
\noindent\textbf{Attention mechanisms:} Recently, there has been a great success in using attention in sequence modeling with deep learning. Recurrent neural networks, long short-term memory \cite{lstm}, and gated recurrent units \cite{gru} are known methods of handling sequential data such as video, text, and speech. For a long time, these were held as state-of-the-art methods but not without flaws. In particular, while these methods can successfully represent dependencies when they are close across input/output sequences, their performance suffers when representing distant relationships. Self-attention is an alternative way to tackle that issue, being capable of modeling relationships among elements in different positions of an input sequence while creating a rich representation. This characteristic of self-attention methods brings a clear advantage over traditional sequence handling, and the Transformer \cite{attentionisall} was the first method to show that it is possible to discard convolutions completely. Transformer networks rely only on self-attention to model representations of input/output sequences, representing the very first choice in more recent NLP applications (usually sequential by nature). Similarly, in image-related tasks, the use of context-aware mechanisms like attention has been shown to generally improve the accuracy of deep neural networks. There is a particular interest in techniques that can be used with minimal effort to improve existing architectures. The Squeeze-and-Excitation (SE) \cite{squeeze-excitation} blocks are drop-in components that model contextual dependencies in channel-wise relationships in feed-forward networks. The Bottleneck Attention Module (BAM) \cite{bam} and the Convolutional Block Attention Module (CBAM) \cite{cbam} are these types of components that propose to do the similar job, additionally integrating both spatial- and channel-wise relationships.

In \cite{aaconv2d}, the principle of multi-headed self-attention from the Transformer network is adapted for 2D inputs, presenting a hybrid layer with attention and convolution operations performed in parallel. These are shown to be compatible with current established deep network architectures, being able to completely replace regular convolutional layers. Unlike BAM and CBAM, which refine existing convolutional feature maps with attention, self-attention augmented convolutions create new attention maps to be fused with their convolutional counterparts. This allows for the network to create more spatially-aware representations, potentially presenting accuracy gains. 


\subsection{Contributions}

In this paper, we introduce a ResNet-inspired \cite{resnet} network, dubbed \textbf{A}ttention-augmented \textbf{Res}Net (ARes-14), conceived upon a self-attention-based mechanism as proposed by \cite{aaconv2d}. ARes-14 was intuitively driven to improve appearance-based gaze estimation, which needs spatial awareness but does not require very deep architectures to be effective (indeed, it only uses 14 layers, as the name suggests).

\begin{table}[t]
\resizebox{\textwidth}{!}{
\begin{tabular}{lcccccc}
\hline
Method  & \begin{tabular}[c]{@{}l@{}}3D gaze\\ 
output\end{tabular} & \begin{tabular}[c]{@{}l@{}}Full-face \\ 
as input\end{tabular} & \begin{tabular}[c]{@{}l@{}}Eye as \\ 
input\end{tabular} & \begin{tabular}[c]{@{}l@{}}Multimodal \\ 
inputs\end{tabular} & \begin{tabular}[c]{@{}l@{}}Spatial \\ 
awareness\end{tabular} & \begin{tabular}[c]{@{}l@{}}Attention \\ 
augmented\end{tabular} \\ \hline\hline
MPIIGaze \cite{mpiigaze} & \checkmark  & -- & \checkmark & -- & -- & -- \\
iTracker \cite{itracker} & -- & \checkmark & \checkmark & \checkmark & -- & --\\
Spatial Weights \cite{writtenallover} & \checkmark & \checkmark & -- & -- & \checkmark & --\\
RT-Gene \cite{rtgene} & \checkmark & \checkmark & \checkmark & -- & -- & --\\
Recurrent CNN \cite{rnngaze} & \checkmark & \checkmark & \checkmark & \checkmark & -- & --\\
Dilated Net \cite{dilatednet} & \checkmark & \checkmark & \checkmark & -- & -- & --\\
FAR-Net \cite{assymetry} & \checkmark & \checkmark & \checkmark & -- & -- & -- \\ \hline\hline
\textbf{Ours} & \textbf{\checkmark} & \textbf{\checkmark} & \textbf{\checkmark} & -- & \textbf{\checkmark} & \textbf{\checkmark} \\ \hline
\end{tabular}
}
\caption{Summary of the state-of-the-art on appearance-based gaze estimation.}
\label{table:methodcomp}
\end{table}

To effectively provide gaze estimation from a monocular camera, we also propose a framework called ARes-gaze, which is comprised of two ARes-14 networks that act as twin feature extractors, taking as inputs full-face images and isolated eye-patches. We showed that, as reported in \cite{aaconv2d} for classification tasks, some of the weights of early attention maps can learn to highlight geometric structures from the full-face images, leading us to hypothesize that self-attention augmented convolutions can fulfill a similar role to the spatial importance maps conceptualized in \cite{writtenallover}. This ability can help the network better focus on facial regions relevant to gaze estimation.

ARes-gaze achieved state-of-the-art results on two challenging data sets. When compared with similar methods of appearance-based gaze estimation, we found a decrease in the average angular error by 2.38\% on the MPIIFaceGaze data set, achieving the second-best result on the EyeDiap data set. Table \ref{table:methodcomp} summarizes the characteristics of our framework in comparison with other state-of-the-art works.

\section{Gaze estimation with self-attention augmented convolutions}

Before going into details of our proposed framework, we review important concepts of appearance-based gaze estimation and the motivation behind the use of attention-augmented convolutional layers.

\subsection{Gaze vector}

3D appearance-based gaze estimation can be comprehended as to find a function capable of mapping an input image, \textbf{\textit{I}}, to a gaze vector, \textbf{\^{g}}. Given that the gaze direction is usually also dependent on head pose, (\textbf{h}), we include this latter into the formulation, thus generically obtaining:

\begin{equation}
    \textbf{\^{g}} = f(\textbf{I},\textbf{h}) \, ,
\end{equation}
where \textbf{\^{g}} is a 2D unit vector with the origin being in the middle point between the subject's eyes. The components that form \textbf{\^{g}} are the pitch ($\mathrm{\hat{\textbf{g}}}_{\theta}$) and yaw ($\mathrm{\hat{\textbf{g}}}_{\phi}$) angles. Here, the mapping function is the proposed trained neural network, and \textbf{h} is implicitly inferred from full-face images. We can then rewrite the generic appearance based formula as $\textbf{\^{g}} = f(\mathbf{I}^{eyes},\mathbf{I}^{face})$.

\subsection{Attention-augmented convolutional layer} \label{sec:attaugconv}
 
First proposed as an alternative base layer for classification \cite{aaconv2d}, attention-augmented convolutions (AAConv) extend the multi-head attention concept from the Transformer network \cite{attentionisall} by applying self-attention to 2D arrays. In regular convolution layers, inter-pixel correlation is usually spatially constrained by the convolutional kernel. This limits the degree to which is possible to relate distant sections from an image that could have relevant relationships. 

\begin{figure}[t!]
\includegraphics[width=\textwidth]{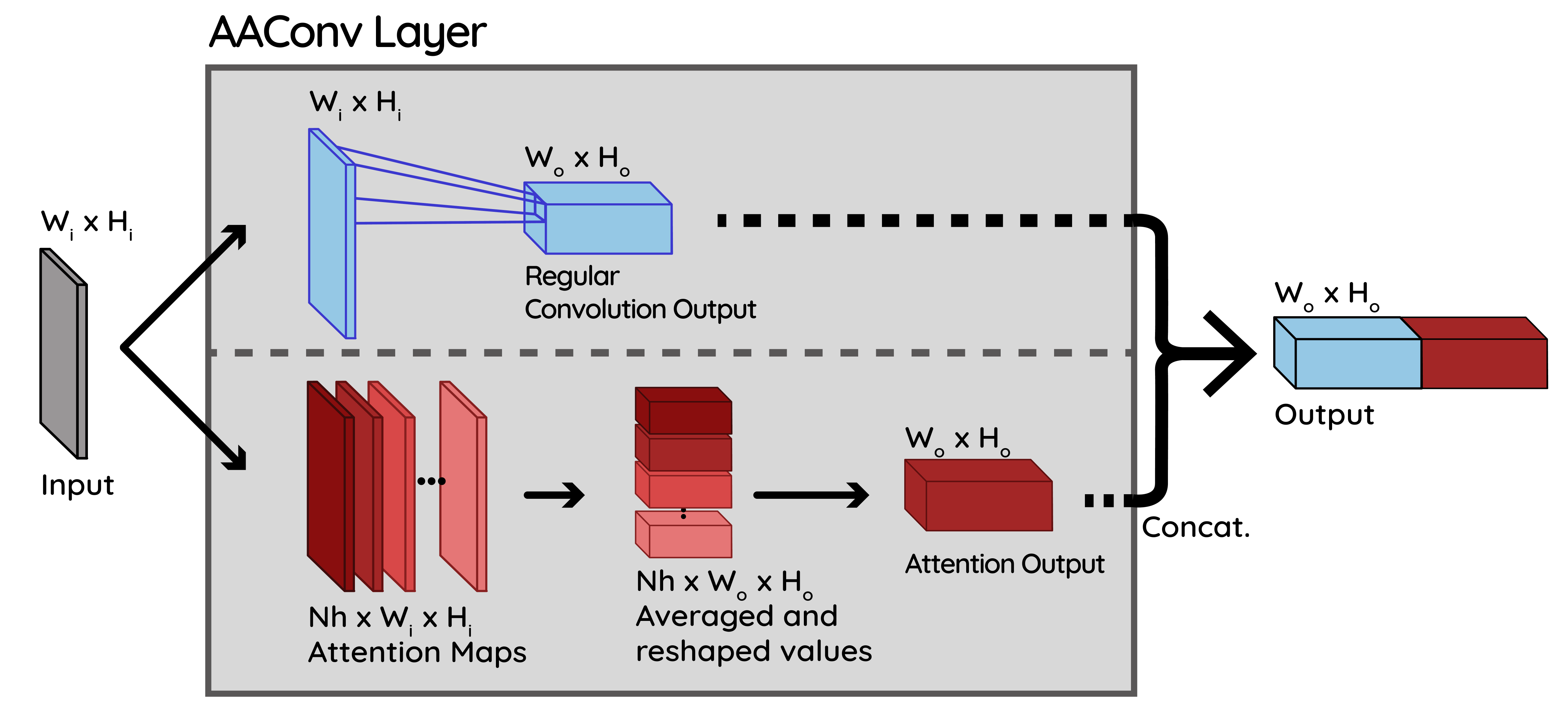}
\caption{Attention augmented convolutional (AAConv) layer as proposed by \cite{aaconv2d}. Nh attention maps are computed for every (w, h) location of the input. The self-attention step consists of performing weighted averaging in these maps, then combining the results by concatenating, reshaping, and mixing ($1 \times 1$ convolution). The resulting feature map is then concatenated with a feature map obtained from a regular convolution performed directly on the input, producing the final output. $W_i$, $H_i$ and $W_o$, $H_o$ are the width and height of the input and output maps, respectively, with $W_o < W_i$ and $H_o < H_i$.}
\label{fig:aaconvlayer}
\end{figure}

Similar to what is done in Transformer networks with 1D sequences, attention-augmented convolutions use self-attention to handle pixel matrices, as depicted in Fig. \ref{fig:aaconvlayer}. Each pass through an AAConv layer can be split into two main parts: The first one through a regular convolutional layer, while the second through a multi-headed attention layer. The outputs ($W_o, H_o$) of each individual attention-head are concatenated and projected onto the original spatial dimensions of height and width of the input ($W_i, H_i$). Additionally, relative positional embeddings \cite{relativepositionembeddings} are expanded to two dimensions in order to encode spatially-relevant information while maintaining translation equivalence \cite{aaconv2d}. In the end, the results from both passes of the convolutional and the multi-headed attention layers are concatenated, forming spatially-aware convolutional feature maps from the input image.  

Expanding the neural network principle of long-distant spatial relationships, it is possible to achieve a clear positive effect when applied to straight-forward classification tasks across a range of different architectures \cite{aaconv2d}. Particularly, we explore if and how to use these concepts as a building-block of CNNs to improve accuracy on a regression task of gaze estimation.

\subsection{ARes-14: A self-attention augmented convolutional backbone}

\begin{figure}[]
\includegraphics[width=\textwidth]{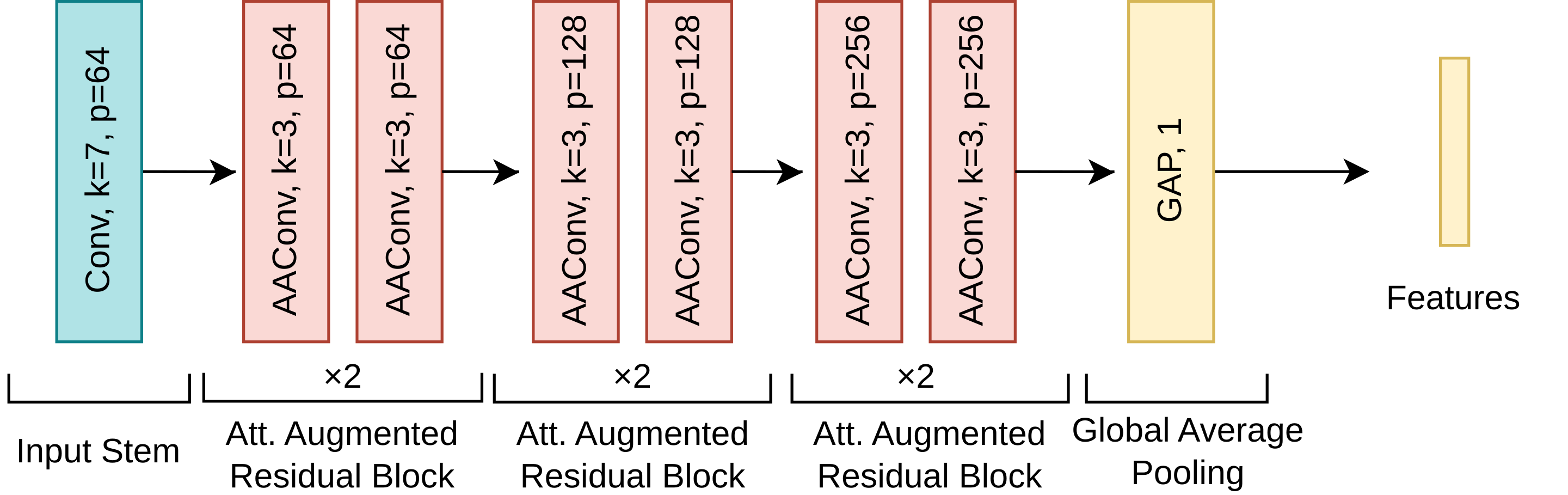}
\caption{ARes-14: Self-attention augmented ResNet with 14 layers. All convolutions in residual blocks are augmented with self-attention, while the input stem remains with conventional convolutions.}
\label{fig:aares14}
\end{figure}

In appearance-based gaze estimation performed by CNNs, shallow networks can be sufficient as long as the task is performed in relatively constrained conditions \cite{rtgene} (across a limited range of head pose and with short distances between subject and camera). These conditions are intrinsic to some available data sets, although they are not a reasonable expectation for in-the-wild applications. These constraints can be simulated in more challenging data by applying preprocessing and normalization procedures (see Section \ref{sec:preprocessing} for more details). The use of these strategies allows us to train with more structured data. Also, procedures like those should still perform well in more complex environments by normalizing the input data before sending it through the prediction network during inference time. The use of shallower networks is of particular importance given the significant computational overhead of training with self-attention in convolutional networks (see \cite{aaconv2d} for a more detailed discussion).

ResNet is a widespread and well understood general-purpose CNN, turning it onto an ideal candidate for a baseline comparison against self-attention augmentation. We started with the shallow version, ResNet-18, and replaced every convolutional layer for an equivalent self-attention augmented convolution with compatible dimensions. The number of parameters was further reduced by removing the last-layer block, essentially shrinking the architecture to 14 layers. Each convolution and AAConv is followed by a batch normalization and activation (ReLU) operation. The ratio between attention channels and output filters (\textit{k}), as well as, the ratio between the key depth and output filters (\textit{v}) were both fixed to 0.25 for every self-attention augmented convolution. Unless otherwise specified, the number of attention heads, $Nh$, is fixed to 8. We called this novel network architecture as \textbf{ARes-14}, which is used as the backbone in our proposed framework for gaze estimation. Figure \ref{fig:aares14} depicts ARes-14 architecture.

\subsection{ARes-gaze: A framework for gaze estimation} \label{sec:aresgaze}

\begin{figure}[]    
\includegraphics[width=\textwidth]{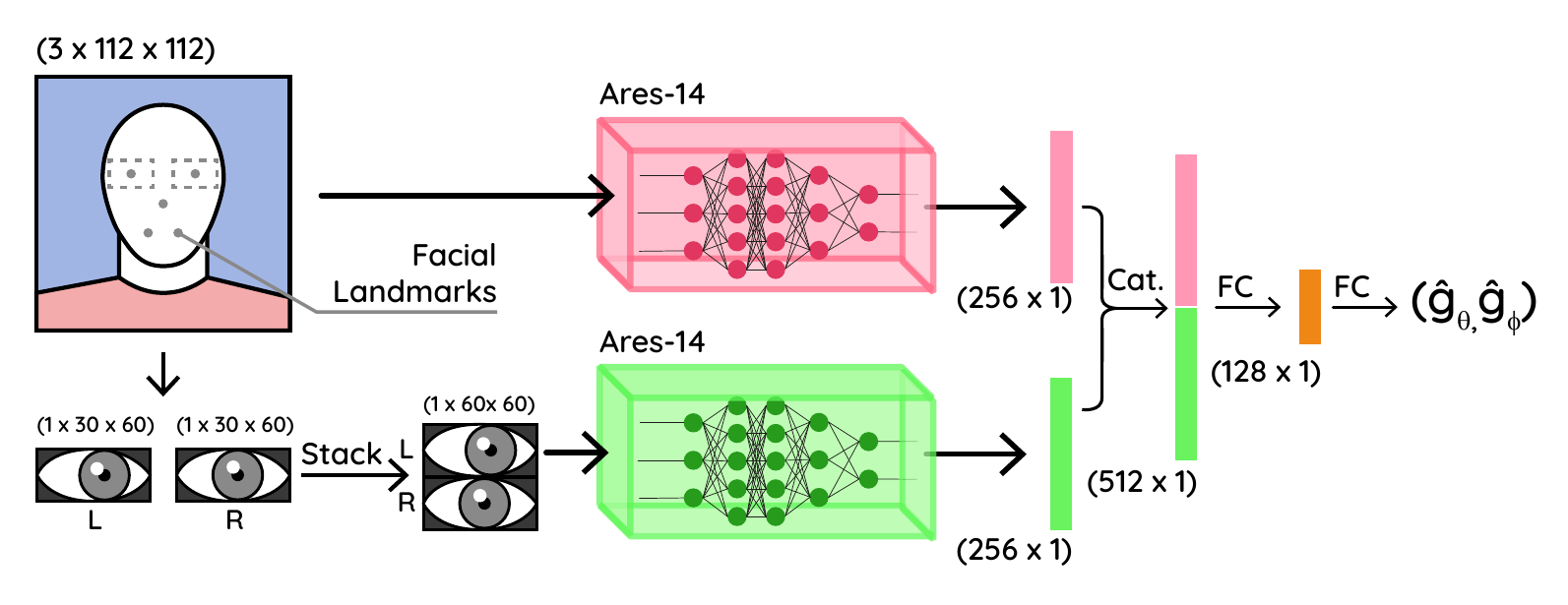}
\caption{ARes-gaze framework. Face- and eye-patches are extracted and separately normalized from the source image, subsequently being sent through twin ARes-14 backbones. The resulting features from each backbone are then concatenated and passed through a prediction stage consisting of two fully-connected layers.}
\label{fig:pipeline}
\end{figure}

To perform gaze estimation, we propose a fairly conventional framework: A two-stemmed network where each branch is an instance of ARes-14, and the extracted features are joined by a shared prediction layer, as shown in Fig. \ref{fig:pipeline}. We used a feature vector of 256 elements obtained from each convolutional backbone after the global average pooling layer, resulting in 512 features to be sent through the prediction layers (see Fig. \ref{fig:aares14}).

Many works have used multi-input frameworks in appearance-based gaze estimation \cite{assymetry, rtgene, dilatednet, rnngaze, geometryconst} since the gaze direction of a subject relies heavily on more than one factor (eyes, head pose, and location, distance, etc). Here our inputs are RGB-face images, normalized for pose and distance, and grayscale eye images, histogram-normalized. 

To extract information from the isolated eye-patches, while some published methods with similar topologies use two networks (one for each eye) \cite{dilatednet, rtgene} or a single network with shared weights (making separate passes for each input) \cite{assymetry}, we employed a single-pass, single-network strategy for the eye branch by stacking the left- and right-eye regions, creating a $1:1$ ratio square input. We study the practical implications of the use of this method in comparison with the other mentioned works in Section \ref{sec:stackedeyeablation}. The extracted-feature vectors from the face and the eyes are then joined by concatenation and passed through a prediction block to output the two values of our gaze vector prediction.

\section{Materials and methods}

\begin{table}[t]
\centering
\begin{tabular}{l|c|c}
\hline
\multirow{2}{*}{\textbf{Characteristics}}  & \multicolumn{2}{c}{\textbf{Data sets}}  \\ \cline{2-3} 
  & \multicolumn{1}{c|}{MPIIFaceGaze \cite{mpiigaze,writtenallover}} & \multicolumn{1}{c}{EyeDiap \cite{eyediap}} \\ \hline\hline

Size  & 45,000 images  & 94 videos  \\\hline 
Image type  & RGB  & RGB-D  \\\hline
Subjects  & 15  & 16  \\\hline
Subject distance  & 40 - 60cm  & 80 - 120cm  \\\hline
Normalized  & \checkmark  & --  \\\hline
Head-pose annotation  & \checkmark  & \checkmark  \\\hline 
Extreme-head pose  & --  & \checkmark  \\ \hline
Extreme-lighting variation  & \checkmark  & --  \\\hline 
Eye-position annotation  & \checkmark  & \checkmark  \\\hline
\end{tabular}
\caption{Comparison of the relevant characteristics of both data sets used in the experiments.}
\label{table:datasets}
\end{table}

\subsection{Training data} \label{sec:trainingdata}

We performed evaluations on the most used publicly-available data sets. Figure \ref{fig:dataset_samples} shows samples of training data of both data sets. Table \ref{table:datasets} summarizes the relevant characteristics of both data sets, described below:  

\begin{figure}[]
\includegraphics[width=\textwidth]{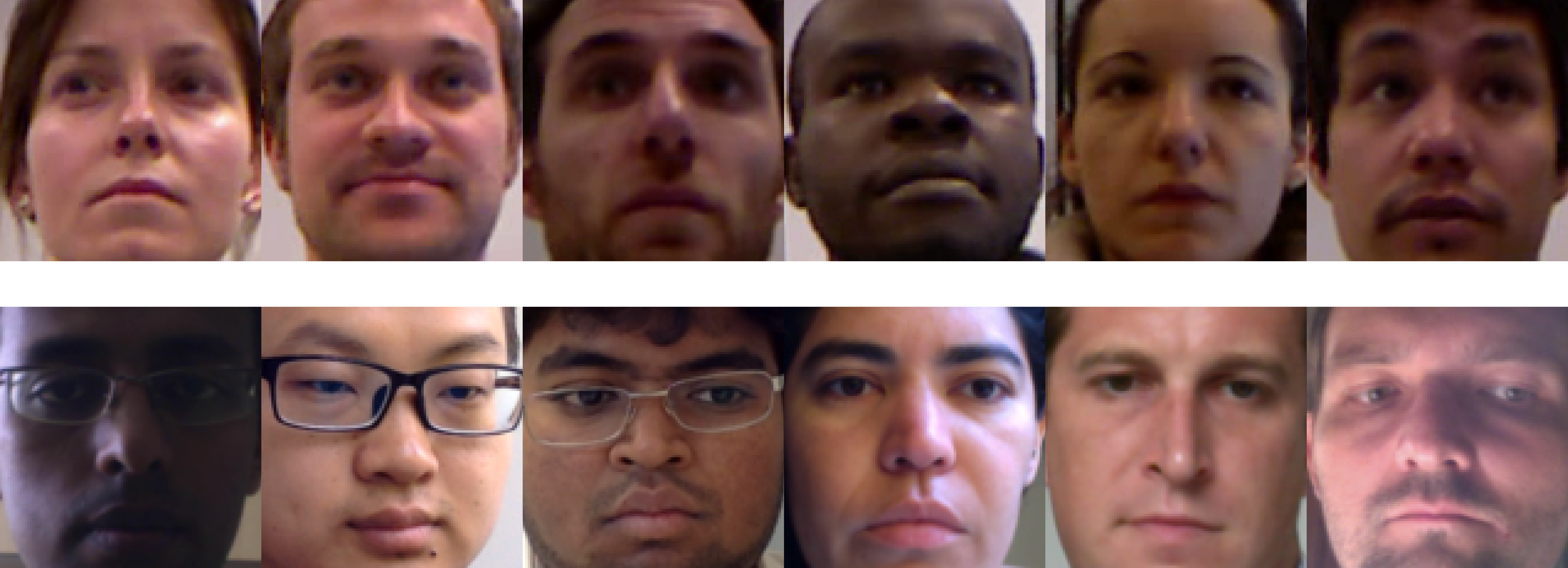}
\caption{Data samples from the EyeDiap \cite{eyediap} (top row) and MPIIFaceGaze \cite{mpiigaze} (bottom row) data sets. The samples on the top row were normalized by following the procedure described in Section \ref{sec:preprocessing}. The bottom samples were taken from the already-normalized MPIIFaceGaze data set. \cite{writtenallover}}
\label{fig:dataset_samples}
\end{figure}

\begin{figure}
    \centering
    \begin{subfigure}[]{0.32\textwidth}
        \includegraphics[width=\textwidth]{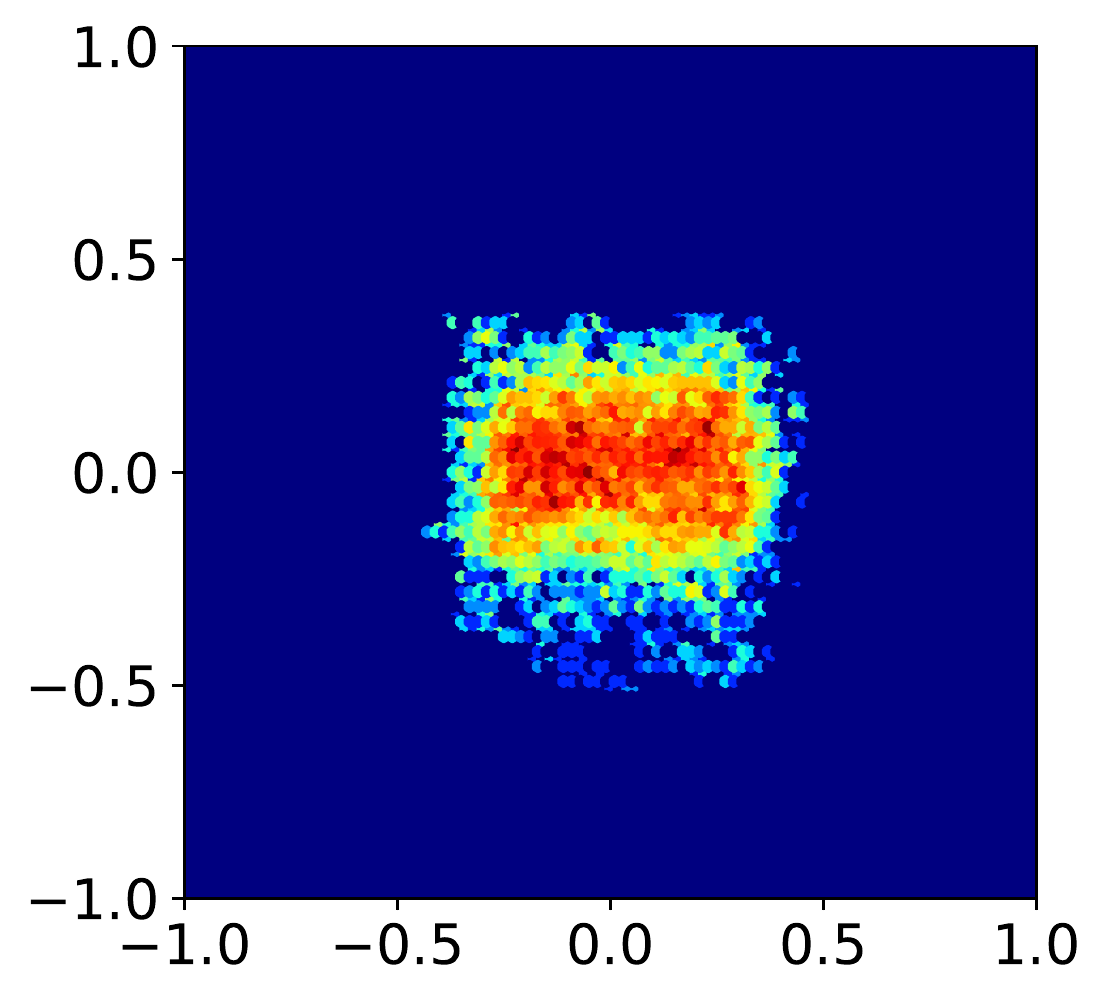}
        \caption{EyeDiap \textbf{g} (static)}
        \label{fig:sub:eyediap_gazedistribS}
    \end{subfigure}
    \begin{subfigure}[]{0.32\textwidth}
        \includegraphics[width=\textwidth]{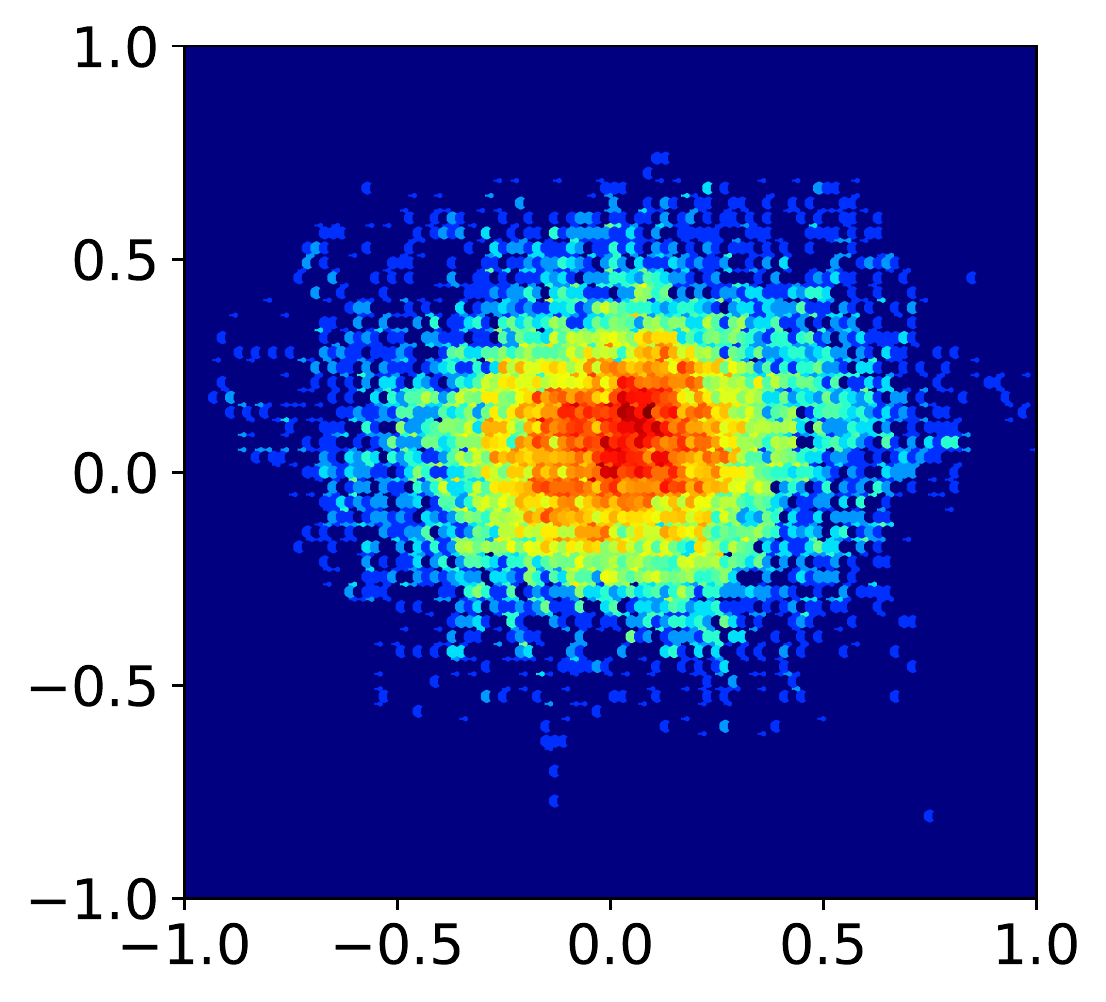}
        \caption{EyeDiap \textbf{g} (mobile)}
        \label{fig:sub:eyediap_gazedistribM}
    \end{subfigure}
    \begin{subfigure}[]{0.32\textwidth}
        \includegraphics[width=\textwidth]{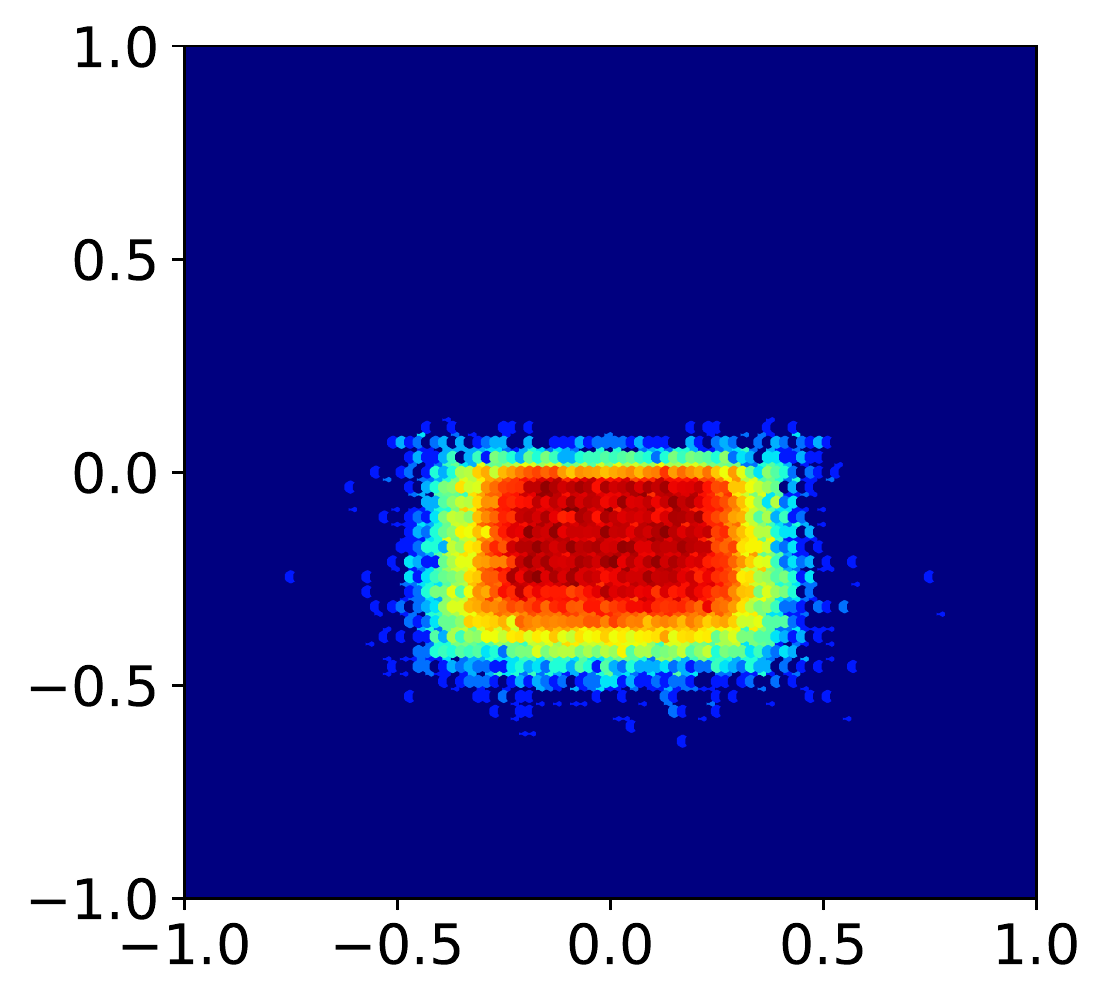}
        \caption{MPIIFaceGaze \textbf{g}}
        \label{fig:sub:mpii_gazedistrib}
    \end{subfigure}
    \newline
    \begin{subfigure}[]{0.32\textwidth}
        \includegraphics[width=\textwidth]{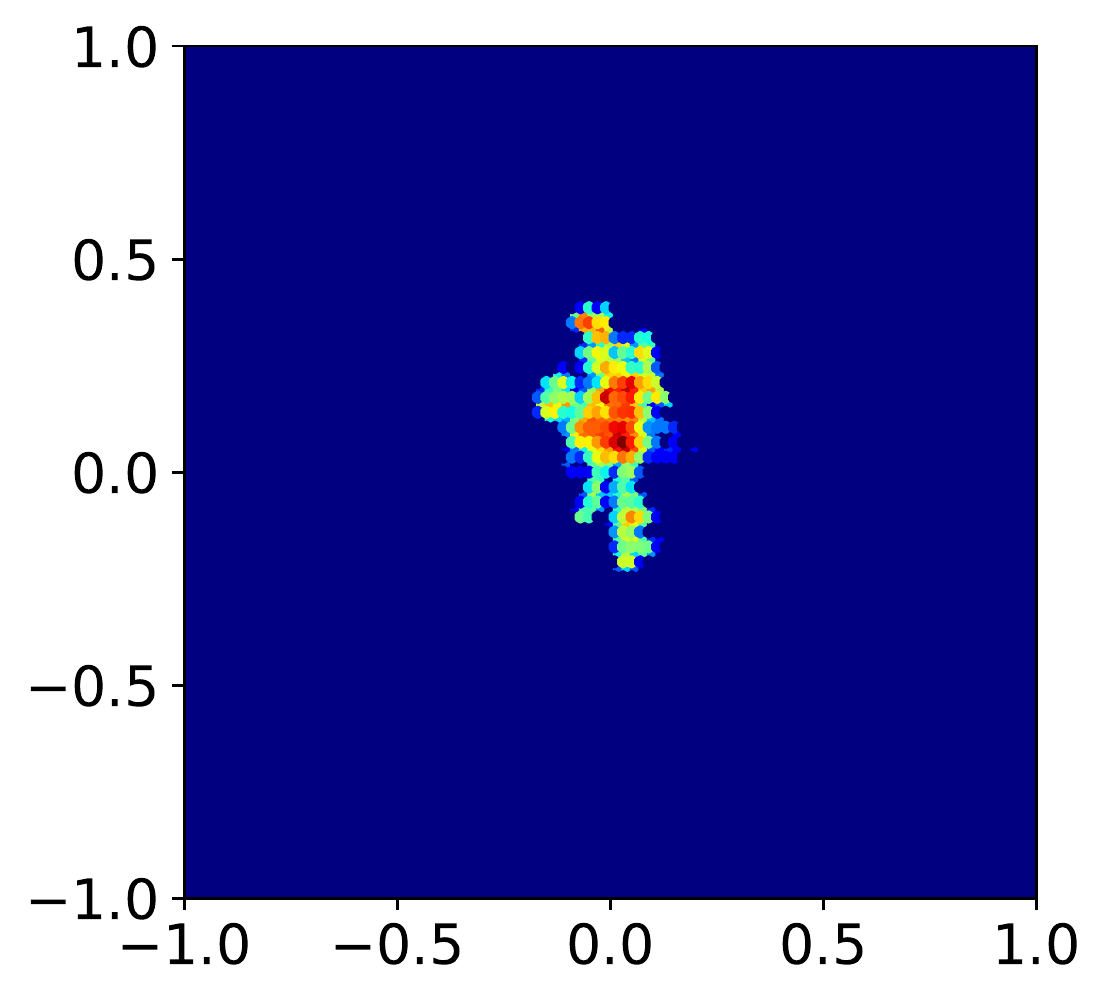}
        \caption{EyeDiap \textbf{h} (static)}
        \label{fig:sub:eyediap_posdistribS}
    \end{subfigure}
    \begin{subfigure}[]{0.32\textwidth}
        \includegraphics[width=\textwidth]{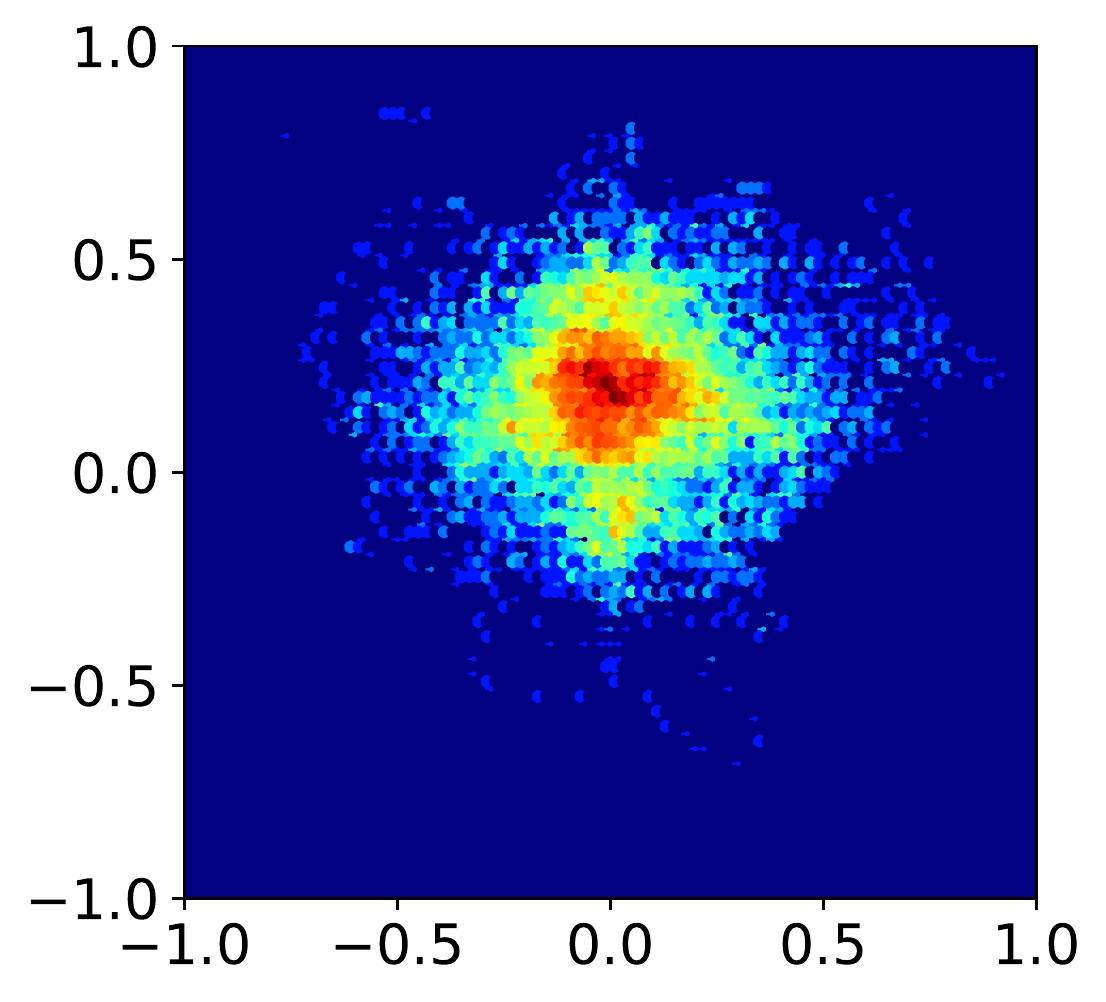}
        \caption{Eyediap \textbf{h} (mobile)}
        \label{fig:sub:eyediap_posdistribM}
    \end{subfigure}
    \begin{subfigure}[]{0.32\textwidth}
        \includegraphics[width=\textwidth]{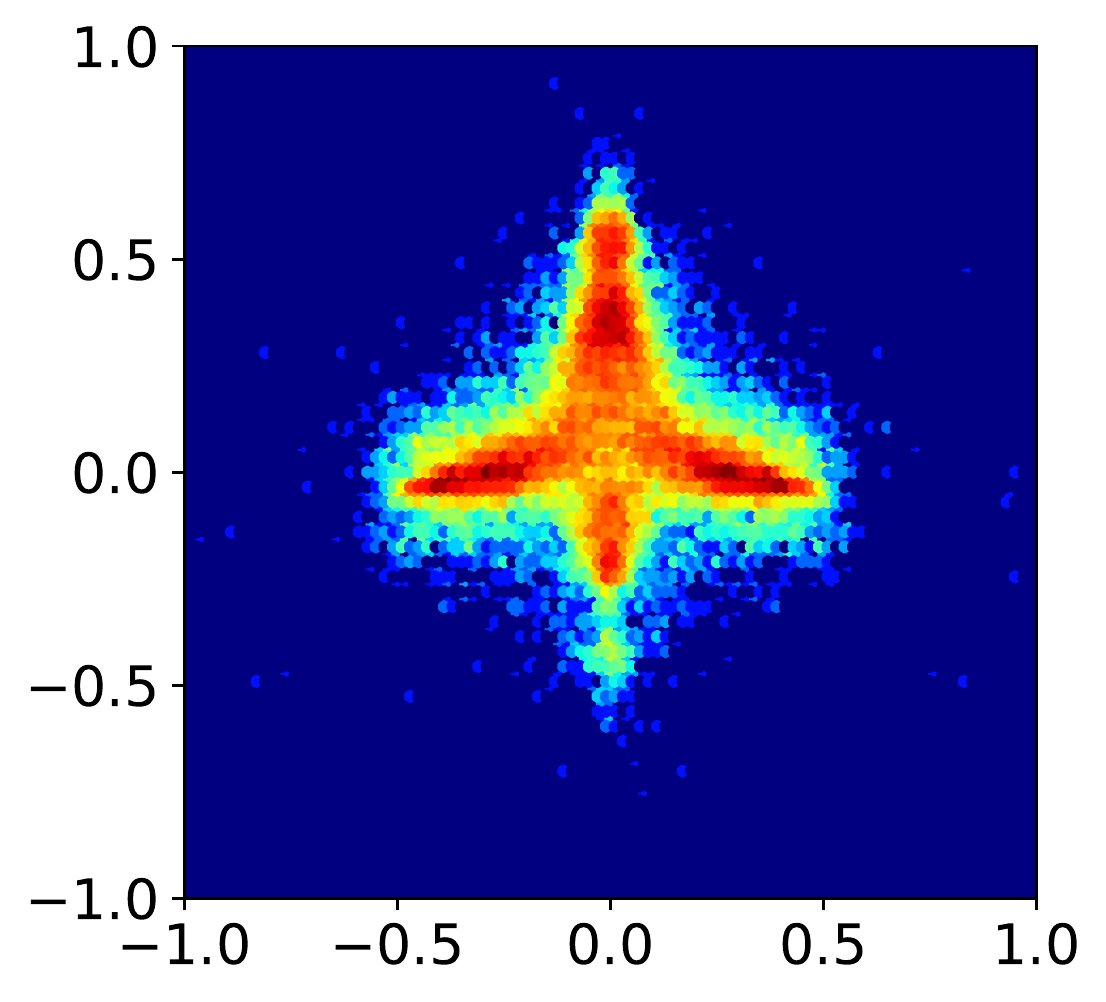}
        \caption{MPIIFaceGaze \textbf{h}}
        \label{fig:sub:mpii_posdistrib}
    \end{subfigure}

\caption{Gaze ground-truth (\textbf{g}) and normalized head pose (\textbf{h}) distribution on the MPIIFaceGaze and EyeDiaps data sets. The latter is split into static or mobile according the subject's head movement. Angle values are displayed in radians.}
\label{fig:gaze_head_distrib}
\end{figure}

\noindent \textbf{The MPIIFaceGaze data set.} The MPIIGaze data set \cite{mpiigaze} was the first to provide unconstrained data for gaze estimation in-the-wild. 15 subjects (9 males, 6 females, and 5 subjects with glasses) were recorded in various sessions during day-to-day use of their laptops, where targets were occasionally displayed at random positions in the screen. The recorded data contains a large number of different conditions of recording locale (inside and outside), illumination, head pose and position, and overall recording quality. Since the original MPIIGaze data set provides cropped-eye regions already, we used its modified version MPIIFaceGaze \cite{writtenallover}, which provides 3,000 full-face, already normalized images for each subject. Figures \ref{fig:sub:mpii_gazedistrib} and \ref{fig:sub:mpii_posdistrib} show the gaze data distribution for the MPIIFaceGaze data set.
\\
\\
\noindent \textbf{The EyeDiap data set} is a collection of 94 videos with 16 different subjects in 3 different modalities: \textbf{Discrete screen target} - where a target was displayed in regular intervals on random locations on a screen, \textbf{continuous screen target} -- in which the target moved along random trajectories in the screen, and \textbf{3D floating target} -- where a small ball was moved along the 3D space between the participant and the camera with the assistance of a thin thread. In our experiments, we used only the modalities where the target was projected onto the screen (continuous and discrete), since, in the floating target-sessions, the small ball would sometimes occlude the subject's face. Two subjects only have video recordings on floating target sessions, so we are left with a total of 14 subjects and 56 videos. For each one of the 14 subjects, there are two different versions of each session: One where the subject's head is fixed and the target is pursued with eye movements only, and another where the subject also moves the head. 

Figures \ref{fig:sub:eyediap_gazedistribS}, \ref{fig:sub:eyediap_gazedistribM}, \ref{fig:sub:eyediap_posdistribS}, and \ref{fig:sub:eyediap_posdistribM} show the gaze and head-pose data distribution for the EyeDiap data set grouped by whether the subject was instructed to move its head (static or mobile). We sampled every $5^{th}$ frame from the recordings to collect the processed data for training. We cleaned the resulting frames by excluding those with missing annotations for the head pose, screen target location, or eye position. Annotations for invalid frames because the subjects have their eyes closed (blinking) or are distracted (looking away from the target) are available for some of the data set's sessions, but not all. For those sessions without such annotations, we manually parsed extracted frames, removing those ones considered invalid. In the end, approximately 44.000 examples remained to carry out with the leave-one-person-out cross-validation. 

\subsection{Data set normalization}\label{sec:preprocessing}

\begin{figure}[]
\includegraphics[width=\textwidth]{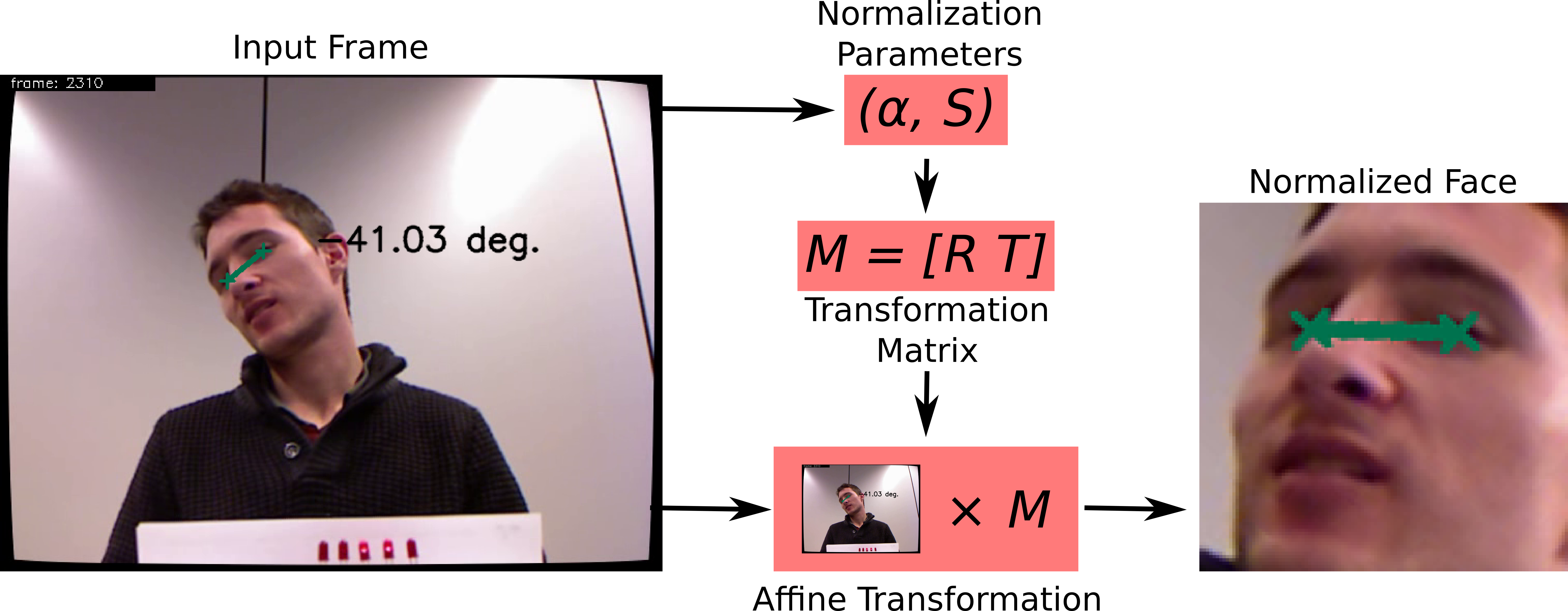}
\caption{Normalization procedure on a sample frame from the EyeDiap data set. The line drawn between the subject's eyes is used to determine the $\alpha$ and $S$ parameters needed to build the transformation matrix $M$.}
\label{fig:normalization_pipeline}
\end{figure}

Similar to \cite{synteshisnormaliz}, we applied an affine transformation to rotate the image as to cancel out the roll-axis angle of the head, and to scale it to the desired size (standardizing the distance from the face to the virtual camera). The effect of that transformation is that relevant facial features are always in the same regions on the input, making the network job easier to recognize patterns in important regions. This procedure is only applied on the EyeDiap \cite{eyediap} data set since the MPIIFaceGaze data set \cite{writtenallover} is already normalized. Figure \ref{fig:normalization_pipeline} illustrates these procedures.

The affine transformation is represented as the matrix $M$, which is defined as $M = [R \ T]$, where $R$ is the rotation component that normalizes the roll-axis rotation, and $T$ is the translation component that ensures the relevant patch from the original frame is in the normalized image. To build $R$, we need the parameters $\alpha$ (rotation) and $S$ (scale). We estimate the angle of the line between the subject's eyes, and use that value as the parameter $\alpha$, which can be obtained with the help of a facial landmark detector, although here we use the annotated position provided by the Eyediap data set. The scale parameter, $S$, controls the distance to the subject in the image. For a squared input, we defined that the distance, $d$, between the left- and right-eye centers should be 40\% of the image width. Given that the face should be vertically centered, this gives us left- and right-eye centers to be $(0.7, 0.35)$ and $(0.3, 0.35)$ on a normalized 0 to 1 scale relative to the input dimensions. 

$S$ can then be given as 
\begin{equation}
    S = \frac{Z * d}{D},
\end{equation}
where $Z$ is the final image size in pixels (considering square images), and $D$ is the original distance between the subject's eye center.

With the angle and scale parameters in hands, we can then build $R$ as follows
\begin{equation}
    R = \begin{bmatrix}
    a  &  b \\
    -b &  a
    \end{bmatrix},
\end{equation}
where a = S * $\cos(\alpha)$, b = S * $\sin(\alpha)$.

$T$ is defined as
\begin{equation}
    T = \begin{bmatrix}
    (1-a) * GO.x  - b * GO.y \\
    b * GO.x + (1-a) * GO.y
    \end{bmatrix} + 
    \begin{bmatrix}
    tX - GO.x \\
    tY - GO.y
    \end{bmatrix},
\end{equation}
where tX = Z * 0.5 , tY = Z * 0.35, and $GO$ is the coordinates of the gaze origin vector (here, defined as the middle point between the left- and right-eye centers). The $tX$ and $tY$ are the correction factors that enable re-positioning the gaze vector in the normalized image.

For the face patch, we used RGB images with an input size of $112 \times 112$ pixels (thus, for our case $Z = 112$). The eye patches are cropped from the normalized face, converted to grayscale, and the histogram normalized before being resized to the input shape of $30 \times 60$. These steps are carried out for both data sets.

\subsection{Implementation Details}\label{sec:training}

The code for the models was written with the PyTorch \cite{pytorch} framework\footnote{Link to git repository will be available after paper acceptance}. All the models were trained for 120 epochs with a batch size of 48 on an HPC cluster equipped with 8 NVidia V100 GPUs. The high computational overhead of training attention-based methods is prohibitive with regards to the batch size, and this needs to be taken into account during the hyperparameter tuning. We used a stochastic gradient descendant (SGD) \cite{sgd} solver with a momentum equal to 0.9, and a weight decay of 0.0003 was empirically found to be optimal in preventing over-fitting. The learning rate is linearly warmed-up for 5\% of the epochs until reaching the value of 0.128, then gradually decreased by cosine annealing \cite{sgdrcosannealing}. The loss used for training the whole model was the smooth $L1$ cost function defined as

\begin{equation}
    \text{smooth}_{L1}(x,y) =
        \begin{cases}
            0.5(x-y)^2,      & \text{if } |x-y| < 1\\
            |x-y| - 0.5,   & \text{otherwise} \, ,
        \end{cases}
\end{equation}
where $x$ and $y$ are the ground-truth and predicted vectors, respectively.

\section{Experimental results} \label{sec:experimentalresults}

To characterize our proposed framework properly, a group of experiments was carried out and divided into two main parts: First, a set of ablation studies were performed to assess the impact of the self-attention augmentation modules on aspects of the ARes-gaze architecture. The main goal of this part is to better understand the optimal conditions to apply AAConvs in our framework. In this part, the studies were namely: The input models of the eye images, the performance of ARes-14 \textit{versus} ResNet14, and the number of attention heads for the self-attention augmented layers. In the second part, some of the external factors that directly impact the performance of gaze estimation were analyzed to explore how our proposed framework can deal with them. Two experiments were performed in this part: Head-pose variation and illumination conditions of the input image.

\subsection{Evaluation methodology}\label{sec:evalmethodology}

To improve reproducibility and reduce the effects of subject's dependence on our evaluations, a leave-one-out cross-validation strategy was used across the subjects from each data set. Considering the characteristics of the data sets used in the experiments (see Table \ref{table:datasets}), N models were trained, where N is the number of available subjects in a data set. For each model, a different subject is held out and used for testing. The final result is the average of the evaluations of all models. On the EyeDiap data set, for example, the final scores are the average performance of the 14 trained models on the held-out subject, each time. Similarly, on the MPIIFaceGaze data set, 15 models were trained and their performance scores were averaged into the final results. 

\subsection{Ablation studies} \label{sec:ablationstudies}

The first evaluation here is grounded on different input models for the eye branch. The goal was to compare our proposed single-branch, single-pass vertical stacking scheme (see Section \ref{sec:aresgaze}) with other strategies adopted by similar methods. The second study was to untangle the effect of self-attention augmentation on different inputs (face and eyes) and the network schemes. The aim is to understand how and where self-attention is effective on the task of gaze estimation, and how it ultimately impacts the performance of the proposed framework. By evaluating isolated portions of the proposed framework with and without self-attention augmentation, these experiments are useful in generating insights on how ARes-14 can be best applied, guiding future researches. Finally, we evaluate the effect of choosing different numbers of attention heads for ARes-14. The multi-headed attention mechanism can present significant computational overhead, so an investigation on the trade-off between number of attention heads and evaluation error drives this choice.

\subsubsection{Evaluating different models of the eye images} \label{sec:stackedeyeablation} 

For the eye-patch branch of our network, the input consists of images of both left and right eyes from the subject. Other published works with similar network topologies either need to perform two forward passes \cite{assymetry} or use a dedicated network branch for each eye \cite{dilatednet, rtgene}. We propose the vertical stacking of eye images to obtain a 1:1 input image that can be processed in a single pass.

We evaluate ARes-gaze against the other mentioned models, considering the following parameters: the number of network parameters, approximated floating-point operations (FLOPs), and average angular error. Three models were considered for the eye branch:

\begin{itemize}
    \item Stacked-eyes input \textbf{(SE)};
    \item Double-pass with shared weights \textbf{(DP)};
    \item Separate branches (three-branch pipeline) \textbf{(TB)}.
\end{itemize}

\begin{table}[]
\centering
\begin{tabular}{ccccc}
\multicolumn{1}{c|}{\multirow{2}{*}{Model type}} & \multicolumn{2}{c|}{Average angular error} & \multicolumn{1}{c|}{\multirow{2}{*}{\# Params (M)}} & \multirow{2}{*}{FLOPs (M)} \\ \cline{2-3}
\multicolumn{1}{c|}{}                                 & \multicolumn{1}{c|}{MPIIFaceGaze} & \multicolumn{1}{c|}{EyeDiap} & \multicolumn{1}{c|}{}                               &                           \\ \hline\hline
SE & \textbf{5.40\degree} & \textbf{7.27\degree} & \textbf{2.810} & \textbf{414} \\
DP & 5.54\degree & 7.42\degree & 2.842 & 422 \\
TB & 5.45\degree & 7.36\degree & 5.619 & 422                  
\end{tabular}
\caption{Results on different input models of the eye images. The evaluated parameters considered are: Average angular error on the EyeDiap and MPIIFaceGaze data sets, number of trainable parameters (\textbf{M}illions) and approximate floating operations (FLOPs, also in \textbf{M}illions) for the three evaluated input models.}
\label{table:RE_params}
\end{table}

As summarized in Table \ref{table:RE_params}, although there is arguably only a small difference in the average angular error, the stacked-input model performed better than the other ones on both data sets. Also the stacked-input model presents roughly the same number of trainable parameters of the shared-weights variety and a significantly lower number when compared to the twin-branch network. These results further validate the adoption of the stacked-eye for all subsequent evaluations.

\subsubsection{ARes-14 evaluation}

With the aim of gauging the effect of self-attention augmentation in multiple stages of ARes-gaze, we evaluated and compared multiple models based on the ARes-14 architecture. First, to see how attention affects different types of input, we trained single-branch networks with and without self-attention augmentation in isolated versions with only eye images as inputs, or only face images as inputs. Second, we applied the ARes-gaze and compare models switching between ResNet-14 and ARes-14 backbones for each input branch. The goal is to explore the contrast between fully convolutional features and self-attention augmented features for gaze estimation.

\begin{table}[t]
\centering
    \begin{tabular}{l|c|c|c|c}
    \hline
    \multicolumn{1}{l|}{} & \multicolumn{2}{c|}{\textbf{Input}} & \multicolumn{2}{c}{\textbf{Dataset}}                            \\
    \cline{2-5} 
    \multicolumn{1}{c|}{\multirow{-2}{*}{\textbf{Network type}}} & \multicolumn{1}{c|}{Eyes}                          & \multicolumn{1}{c|}{Face} & \multicolumn{1}{c|}{MPIIFaceGaze} & \multicolumn{1}{c}{EyeDiap} \\ \hline\hline
    
    \rowcolor{customgray} 
    Regular & $\blacksquare$ &  & 5.40\degree & 7.27\degree  \\ \hline
    Attention & $\blacksquare$ &  & 5.33\degree & 6.02\degree \\ \hline
    \rowcolor{customgray} 
    Regular &  & $\blacksquare$ & 4.71\degree & 7.42\degree  \\ \hline
    Attention &  & $\blacksquare$ & 4.46\degree & 6.10 \\ \hline
    \rowcolor{customgray} 
    Regular & $\blacksquare$ & $\blacksquare$  & 4.46\degree  & 6.09\degree \\ \hline
    Regular & $\blacksquare$ &  & \multirow{2}{*}{4.42\degree} & \multirow{2}{*}{5.81\degree} \\
    Attention &  & $\blacksquare$ &  &  \\ \hline
    \rowcolor{customgray} 
    Regular &  & $\blacksquare$ &  &  \\ 
    \rowcolor{customgray} 
    Attention & $\blacksquare$ &  & \multirow{-2}{*}{4.52\degree} & \multirow{-2}{*}{5.84\degree} \\ \hline
    Attention & $\blacksquare$ & $\blacksquare$    & \textbf{4.17\degree}  & \textbf{5.58\degree}  \\ \hline
    \end{tabular}
\caption{Results of attention-augmented versus regular convolutional layers on the backbones of ARes-gaze. Best results are highlighted.}
\label{table:attablation}
\end{table}

The results for each model are laid out in Table \ref{table:attablation}. The network types are comprised of single regular, single attention or both regular/attention branches. For the single-branch networks (with either only face or only eyes as inputs), we observe a drop of more than 17\% on the average angular error on the EyeDiap data set when using self-attention augmented convolutions. When compared with its regular convolutional form, ARes-gaze reduces the average error by 6.5\% on the MPIIFaceGaze data set and by 8.4\% on EyeDiap.

\subsubsection{Determining the number of attention heads}\label{sec:nhablation}

In all evaluations reported in \cite{aaconv2d}, on the use of AAConvs in classification tasks, the accuracy gains are on architectures using a fixed number of attention-heads, specifically $Nh=8$. In this section, we evaluate ARes-gaze considering other values of $Nh$.

Table \ref{table:attheadablation} shows the average angular errors found on the MPIIFaceGaze and EyeDiap data sets. Notably, for the MPIIFaceGaze data set, when using less than 4 attention-heads, the ARes-gaze architecture performs worse than the purely convolutional baseline, with the evaluation error proportionally decreasing with the increase of attention-heads. On the EyeDiap data set, the results follow the same tendency with $Nh = 2$ and $Nh = 4$, which are only marginally better than the baseline network. In both data sets, there is a sudden and significant improvement in the results when $Nh = 8$. It is worth noting that given the high computational overhead of AAConv layers, with the available hardware (as described in Section \ref{sec:training}) it was not viable to perform experiments with $Nh > 8$.

\begin{table}[t]
    \centering
    \begin{tabular}{lll}
    Method      & MPIIFaceGaze & EyeDiap     \\\hline\hline
    Baseline    & 4.46\degree   & 6.09\degree  \\
    ARes-gaze (Nh=2) & 4.93\degree   &  5.98   \\
    ARes-gaze (Nh=4) & 4.36\degree   &  5.99       \\ 
    \textbf{ARes-gaze (Nh=8)} & \textbf{4.17}\degree   & \textbf{5.58}\degree    
    \end{tabular}
\caption{Results of average angular errors on different numbers of attention-heads per attention layer. Best results are highlighted.}
\label{table:attheadablation}
\end{table}

\subsection{Comparison of ARes-gaze with other appearance-based methods}

We selected six appearance-based methods that take as input either full-face images or a combination of full-face images and other inputs. All these methods output a single-gaze vector with origin in the center of the face or in the middle-point of the eye. The selected methods were: the iTracker in its original form \cite{itracker} and with AlexNet backbone \cite{writtenallover}, the CNN with spatial-weights mechanism \cite{writtenallover}, RT-Gene (a version of 4 ensembles with the best reported results) \cite{rtgene}, the CNN with dilated convolutions proposed in \cite{dilatednet}, and the eye-asymmetry based FAR-Net \cite{assymetry}. These are approaches we consider similar to ours, which were compared over the average 3D-angular error on the chosen data sets. Except for RT-Gene and iTracker (AlexNet), which do not report evaluations on the EyeDiap data set, all compared methods use the same or a similar protocol to extract data from the videos, as described in Section \ref{sec:trainingdata}.

The results are reported by considering two versions of our architecture: The full ARes-gaze and ARes-gaze without self-attention augmentation.

\begin{table}[t]
\centering
\begin{tabular}{lcc}
Method               & MPIIFaceGaze & EyeDiap \\ \hline\hline
iTracker \cite{itracker}             & 6.2\degree      & 8.3\degree     \\
iTracker (AlexNet) \cite{itracker, writtenallover}   & 5.6\degree      &    --  \\
Spatial Weights CNN \cite{writtenallover}  & 4.8\degree      & 6.0\degree     \\
RT-Gene (4 Ensemble) \cite{rtgene} & 4.3\degree      & --\degree     \\
Dilated CNN \cite{dilatednet}          & 4.5\degree      & \textbf{5.4\degree}     \\
FAR-Net \cite{assymetry}              & 4.3\degree      & 5.7\degree   \\ \hline
Baseline         & 4.5\degree        &      6.1\degree        \\ 
ARes-gaze (Nh = 8)     & \textbf{4.2\degree}      &        5.6\degree
\end{tabular}
\caption{Results of average angular error compared with other appearance-based methods. Best results are highlighted.}
\label{table:maincomparison}
\end{table}

When compared to the other methods, our ARes-gaze framework with twin ARes-14 backbones reached state-of-the-art results on the MPIIFaceGaze data set, and the second-best place on the EyeDiap data set, being only 0.2 degrees behind the Dilated CNN \cite{dilatednet} (see Table \ref{table:maincomparison}). It is worth noting that no other method was able to have superior results on both data sets.

\subsection{Evaluating external factors in gaze estimation}

In in-the-wild gaze estimation applications, the subject's head pose and external illumination conditions are to be considered unconstrained. As such, these are highly relevant factors to be considered when proposing new approaches for gaze estimation. As discussed in Section \ref{sec:preprocessing}, we applied color-level normalization, enforcing roll-angle normalization during training and inference to reduce the complexity of our model when the data set does not offer already normalized images. Regardless, edge cases of extreme conditions are still challenging, and there are yet the pitch and yaw angles to be concerned. To see how self-attention augmentation in convolutions affects robustness to these factors, we evaluated both our completely attention-augmented model and the traditional convolutional baseline in isolated scenarios.

\subsubsection{Head pose}

    \begin{figure}[]
        \centering
        \begin{subfigure}[]{0.2\textwidth}
            \includegraphics[width=\textwidth]{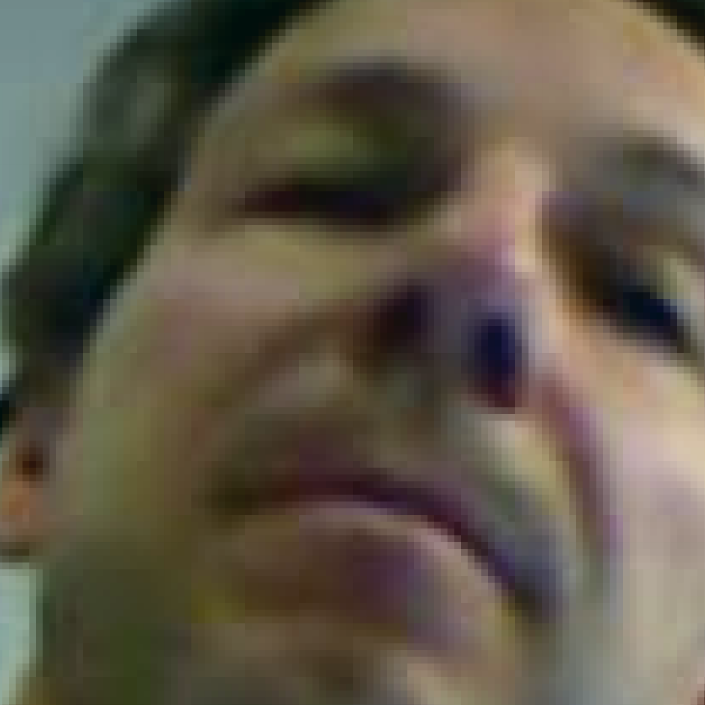}
            \caption{}
            \label{fig:sub:extremeposeeyediap_a}
        \end{subfigure}
        \begin{subfigure}[]{0.2\textwidth}
            \includegraphics[width=\textwidth]{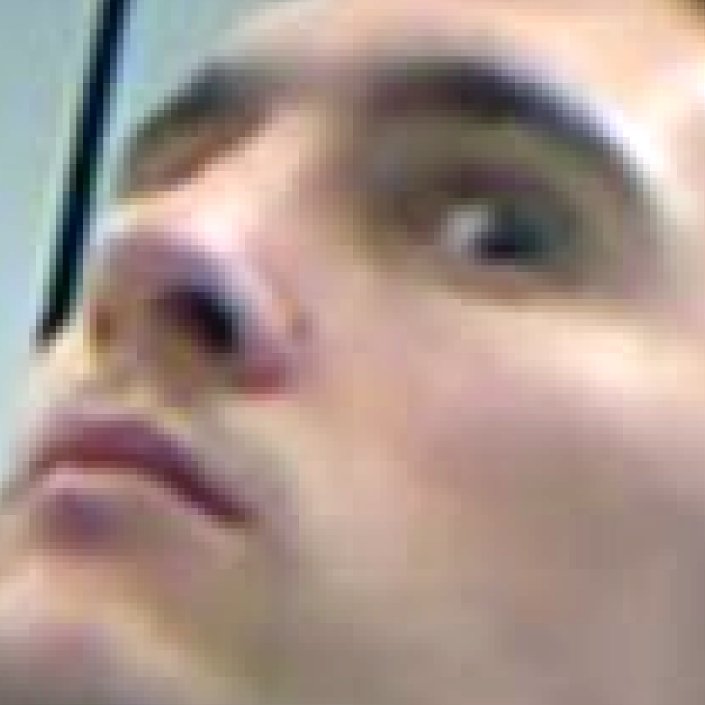}
            \caption{}
            \label{fig:sub:extremeposeeyediap_b}
        \end{subfigure}
        \begin{subfigure}[]{0.2\textwidth}
            \includegraphics[width=\textwidth]{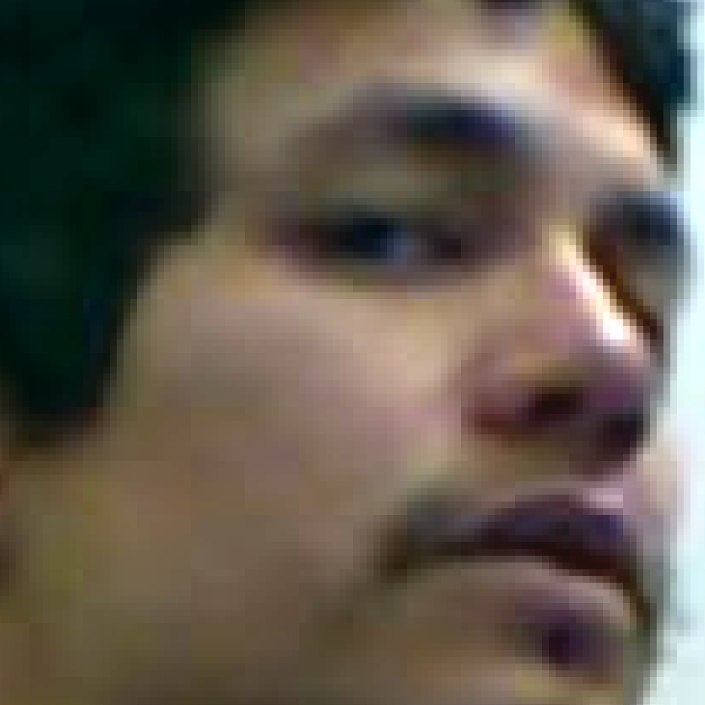}
            \caption{}
            \label{fig:sub:extremeposeeyediap_c}
        \end{subfigure}
        
    \caption{Samples with extreme head pose angles from the EyeDiap dataset.}
    \label{fig:extremeposeeyediap}
    \end{figure}

To evaluate how each model performs to the subject's head pose, we used the EyeDiap data set due to its larger variety (refer to Fig. \ref{fig:gaze_head_distrib}) of head pose and edge cases of extreme pose conditions (see image samples in Fig. \ref{fig:extremeposeeyediap}). We used the data set annotation for head-pose angles to correlate every sample prediction error to the subject's head pose. We then performed 2D binning to obtain the average angular error on intervals of 0.20 radians for pitch and yaw.

\begin{figure}[]
    \centering
    \begin{subfigure}[]{0.49\textwidth}
        \includegraphics[width=\textwidth]{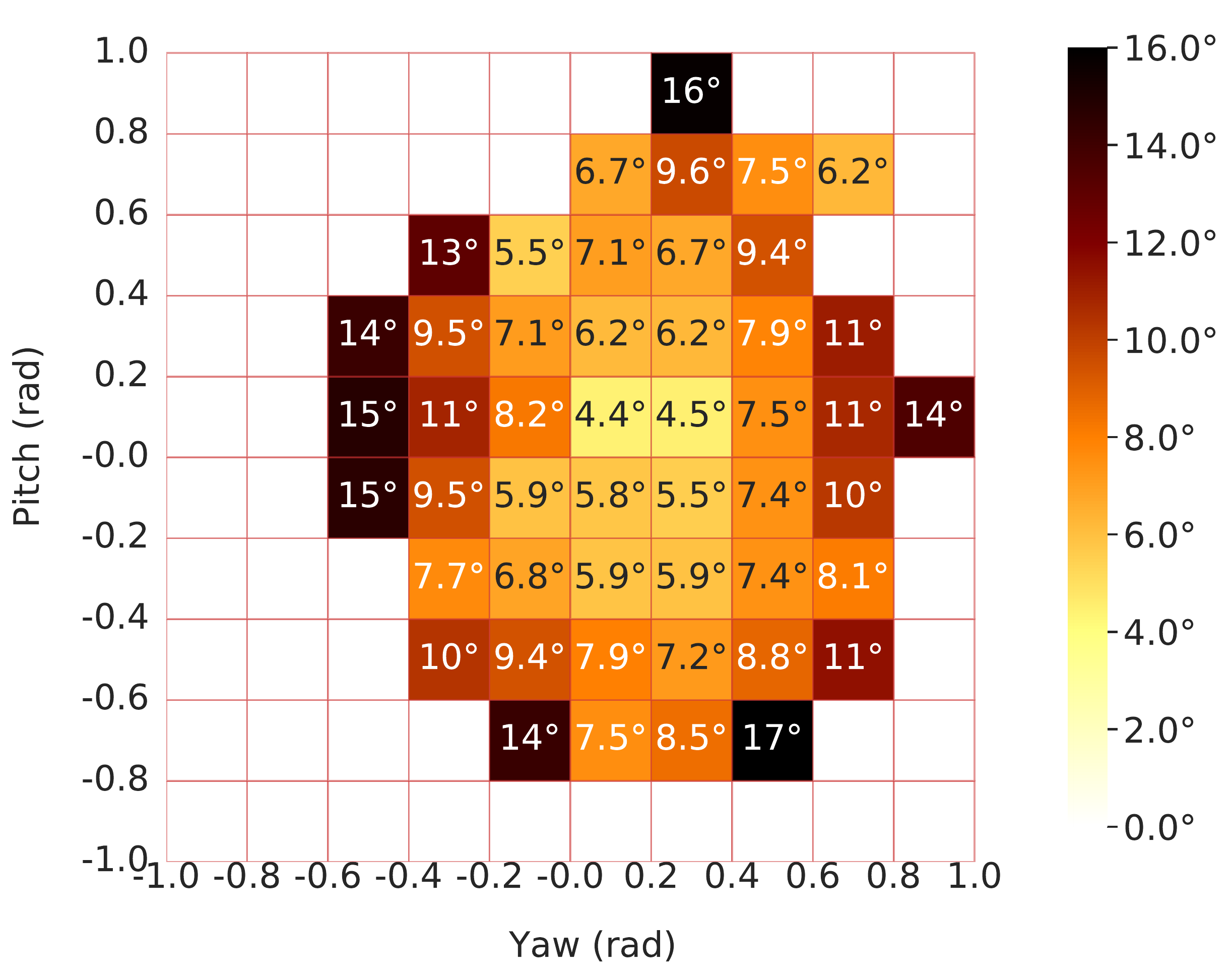}
        \caption{Baseline convolutional model}
        \label{fig:sub:heapose_heatmap_baseline}
    \end{subfigure}%
    \begin{subfigure}[]{0.49\textwidth}
        \includegraphics[width=\textwidth]{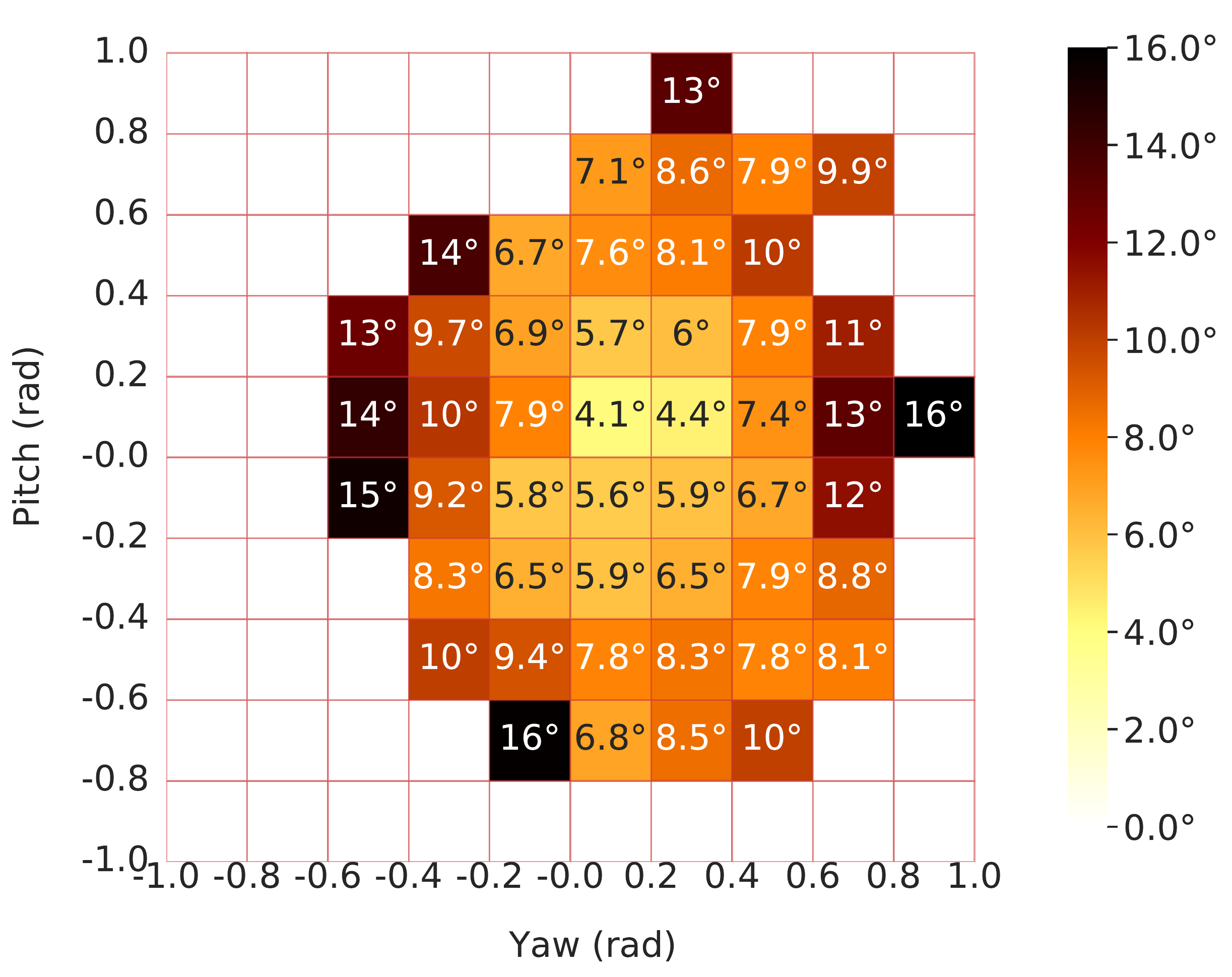}
        \caption{attention-augmented model}
        \label{fig:sub:heapose_heatmap_aaug}
    \end{subfigure}
\caption{Distribution of mean angular error of baseline and attention-augmented models across head poses in the EyeDiap dataset.}
\label{fig:headpose_heatmap}
\end{figure}

\begin{figure}[]
\centering
\includegraphics[width=0.7\textwidth]{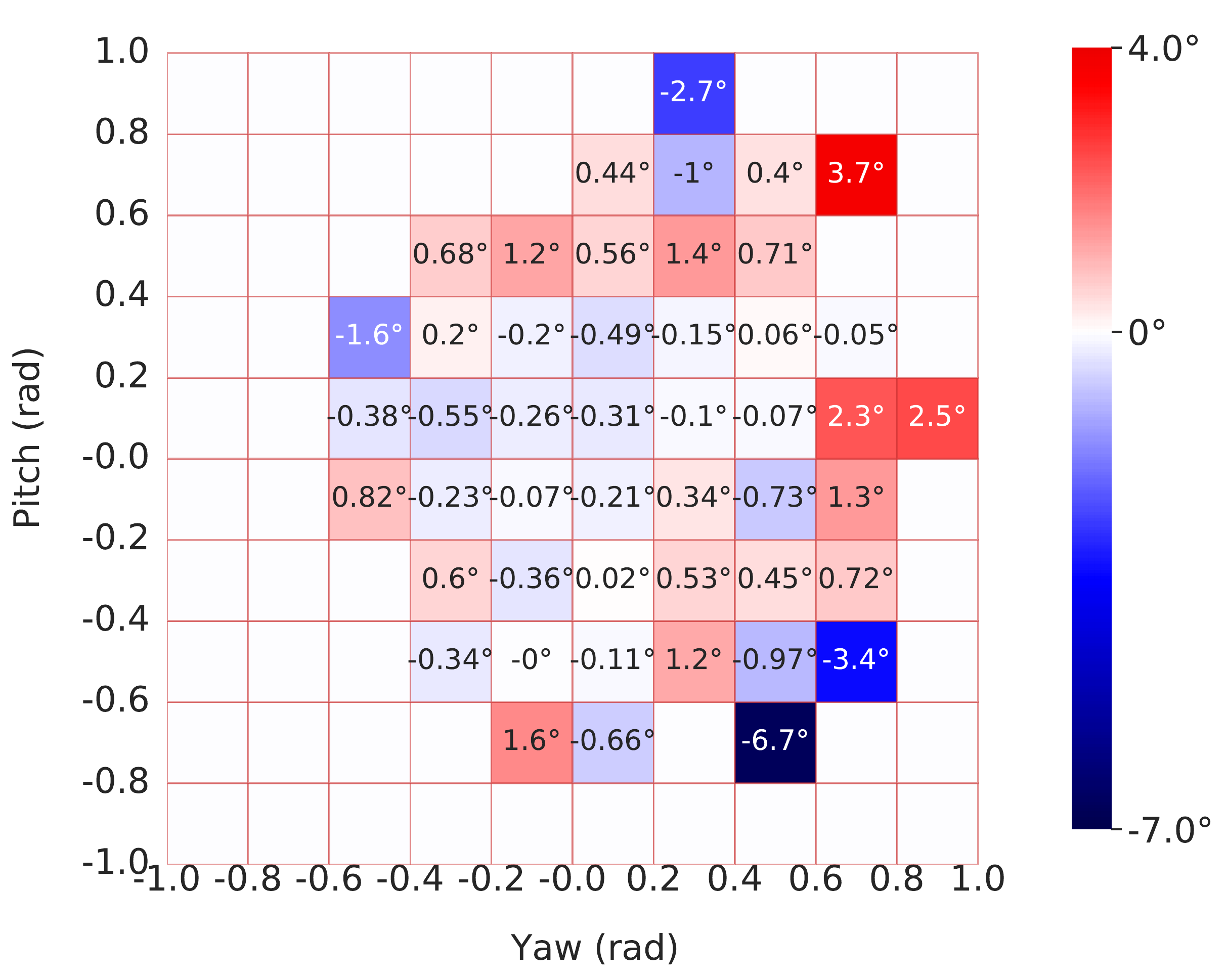}
\caption{Distribution of angle-error difference between attention-augmented and baseline models on the head-pose evaluation. Blue boxes (negative numbers) mean an improvement over the baseline model, or a drop in the average angular error. Similarly, red boxes mean regions where there was an increase in the average angular error.}
\label{fig:diff_headpose_heatmap}
\end{figure}

Figure \ref{fig:headpose_heatmap} shows the results for both self-attention augmented and regular convolutional based architectures. The plots make clear that the overall gains are obtained across most of the pitch and yaw head-pose spectrum. This notion is further reinforced by Fig. \ref{fig:diff_headpose_heatmap} where the average angular error difference between the two plots is plotted. The overall decrease in average error appears mostly uniform outside of the most extreme cases. For those, we observe that the larger gains obtained by the ARes-gaze model were in regions of extreme pitch angle (negative and positive), and the heavier losses were in regions of high yaw angles. It should also be noted that the overall magnitude of the gains in average accuracy is greater than that of eventual losses suffered by the model (color bar in Fig \ref{fig:diff_headpose_heatmap}).

\subsubsection{Illumination conditions}

\begin{figure}[]
\includegraphics[width=\textwidth]{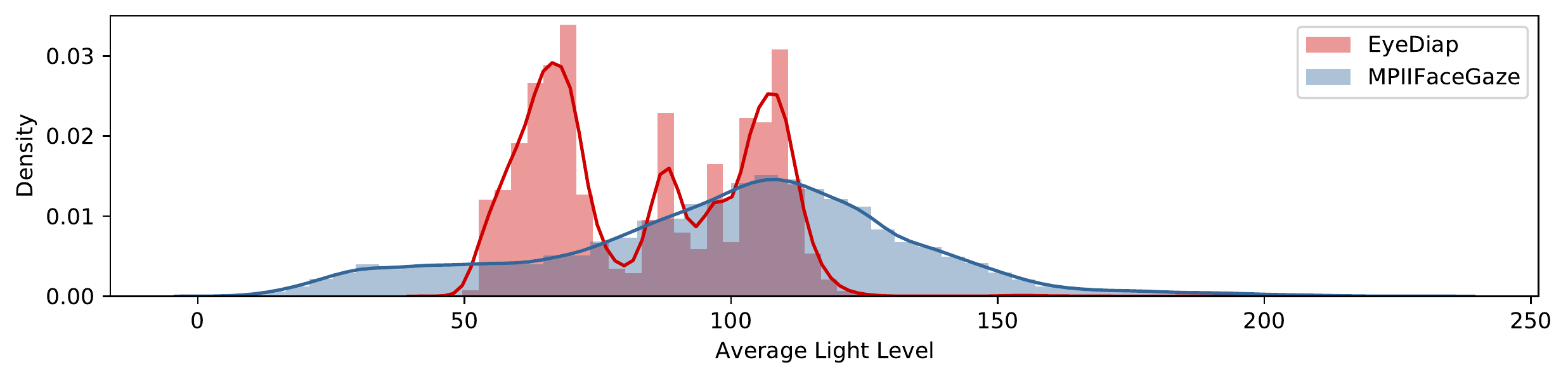}
\caption{Distribution of average light intensity per sample on the MPIIFaceGaze and EyeDiap datasets.} 
\label{fig:lightdistrib_datasets}
\end{figure}

\begin{figure}[t]
    \centering
    \begin{subfigure}[]{0.3\textwidth}
        \includegraphics[width=\textwidth]{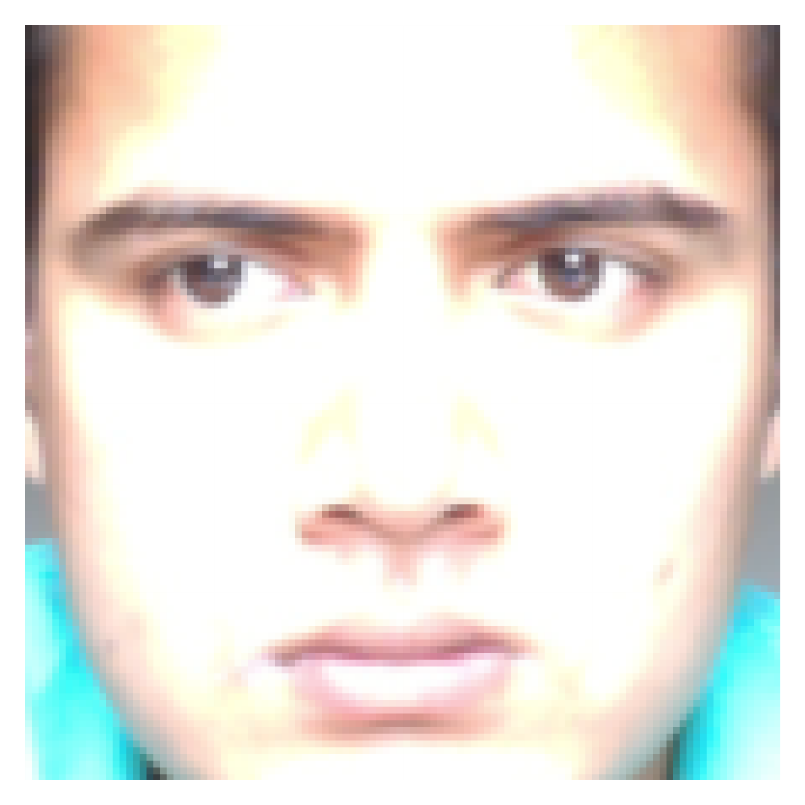}
        \caption{Sample with average light level greater than 220.}
        \label{fig:sub:light_sample_mpii_high}
    \end{subfigure}
    \begin{subfigure}[]{0.3\textwidth}
        \includegraphics[width=\textwidth]{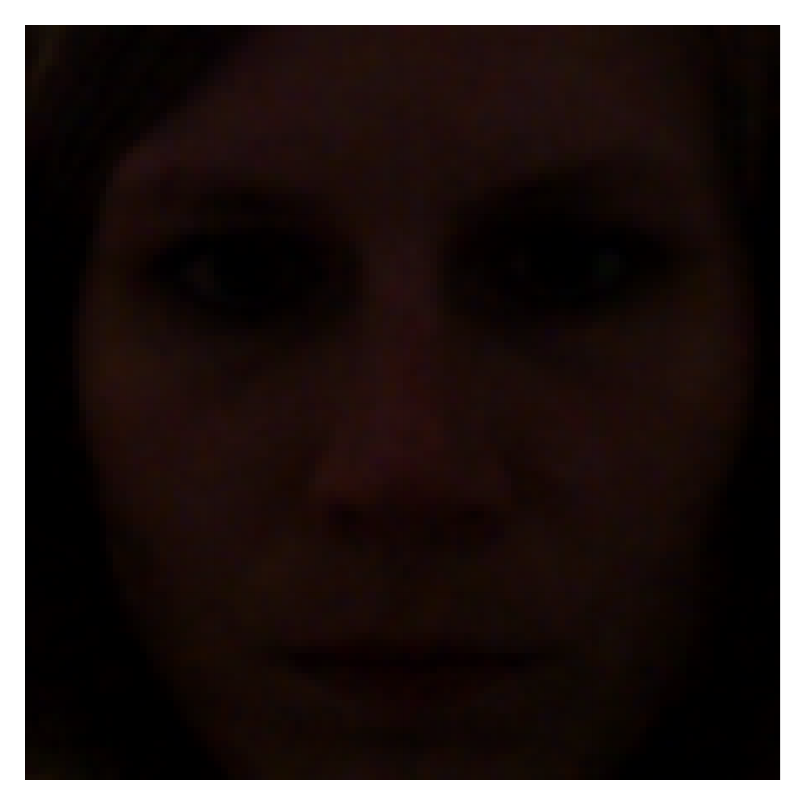}
        \caption{Sample with average light level lower than 20.}
        \label{fig:sub:light_sample_mpii)low}
    \end{subfigure}
\caption{Samples of edge cases w.r.t lighting conditions from the MPIIFaceGaze data set. These examples were randomly sampled from images with average light level greater than 220 and lower than 20.}
\label{fig:light_sample_mpii}
\end{figure}

In Fig. \ref{fig:lightdistrib_datasets}, the illumination distribution is plotted on both the MPIIFaceGaze and EyeDiap data sets. The light level values are obtained per sample by averaging the pixel intensity to obtain a value between 0 (completely dark) and 255 (completely bright). The plot shows that the images from MPIIFaceGaze data set is more varied to lighting conditions (illustrated by the sample edge-case from both extreme light and darkness shown in Fig. \ref{fig:light_sample_mpii}), while the average intensity of image pixels approximate to a normal distribution. These properties should be beneficial to our evaluation, so we chose to analyze the effect of self-attention augmentation \textit{versus} lighting on the MPIIFaceGaze data set. 

 \begin{figure}[]
\begin{center}
\includegraphics[width=\textwidth]{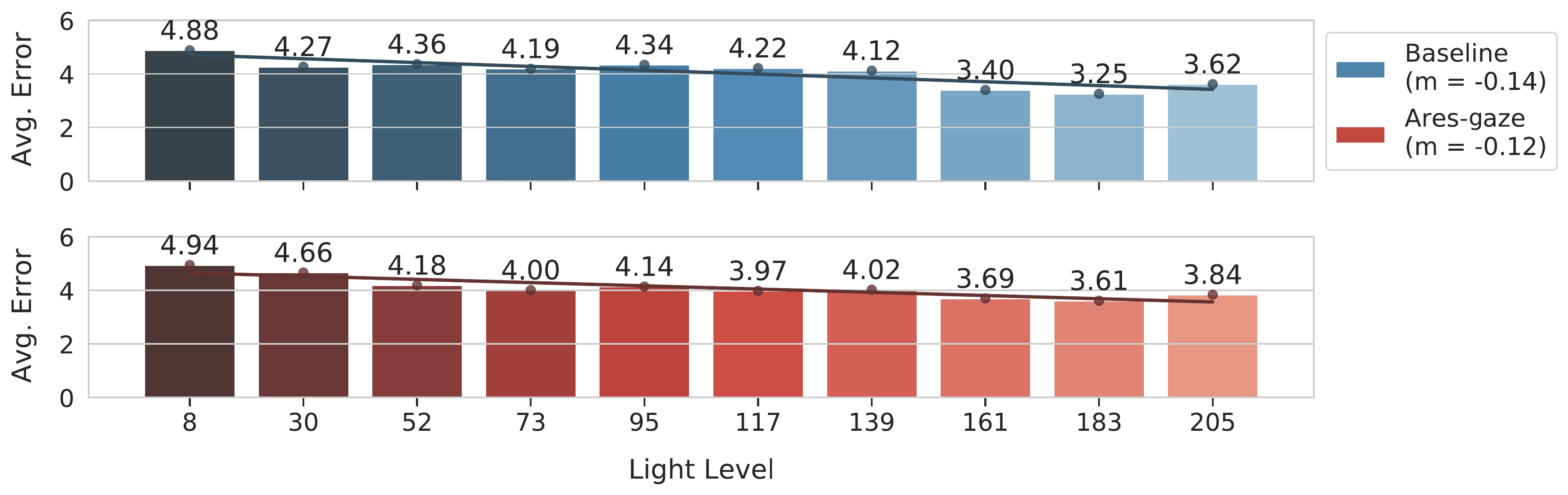}
\end{center}
\caption{Evaluation of model accuracy \textit{versus} lighting conditions of input images. The MPIIFaceGaze data was split into 10 bins with regard to light levels, with the X-axis showing the average level of each bin. The Y-axis is the average angular error in degrees. A regression line drawn across each model bars, with its slope value ($m$) being shown in the plot legend.}
\label{fig:mpii_light_eval}
\end{figure}

Figure \ref{fig:mpii_light_eval} shows an overlapping evaluation of both baseline and ARes-gaze models by light-level intervals. We split the 0-255 light-level range into 10 bins, and the results are averaged across all 15 subjects in relation with these bins. Clearly there is an inverse relationship between light level and angular error that behaves somewhat linearly. Additionally, the last bin, representing extreme high-light levels (overly lit images) shows a small spike in the averaged angular error. This situation reinforces the intuitive notion that appearance-based gaze estimation models have worse accuracy with both poorly lit and overexposed input images.

To quantify the sensibility of each model to lighting conditions, we fit a regression line across the angle error of each bin, and calculated its slope ($m$). The closer to zero the slope is, the lower is the model sensibility to light. This experiment showed that ARes-gaze had a slightly smaller slope inclination, although the difference was not enough to justify conclusions about its robustness to lighting conditions in comparison with the purely convolutional baseline.

\subsection{Result analysis}

\subsubsection{On the use of self-attention augmented convolutions for gaze estimation}

First, we evaluated the difference between using eye images \textit{versus} using the entire face as inputs. In these experiments, the eye images were stacked before being forwarded through the network, as reported in Section \ref{sec:stackedeyeablation}. Intuitively, the difference between using full-face images and isolated-eye regions as inputs is the scope of the information that the network is able to extract. With full-face images, CNN has the chance to learn not only from the eyes themselves but also extract head-pose information from regions such as the nose and mouth. This comes with the drawback of the subject's eyes having a lower resolution, thus limiting the amount of information present in their regions. In contrast, using isolated eye-patches should allow the network to extract more detailed information about the pupils' positions, turning the network to be more sensitive to smaller changes in the eye movement. In this case, the drawback is the absence of elements that can inform the network about the subject's head pose, which has relevance to the final prediction.

\begin{figure}[]
\begin{center}
\includegraphics[width=\textwidth]{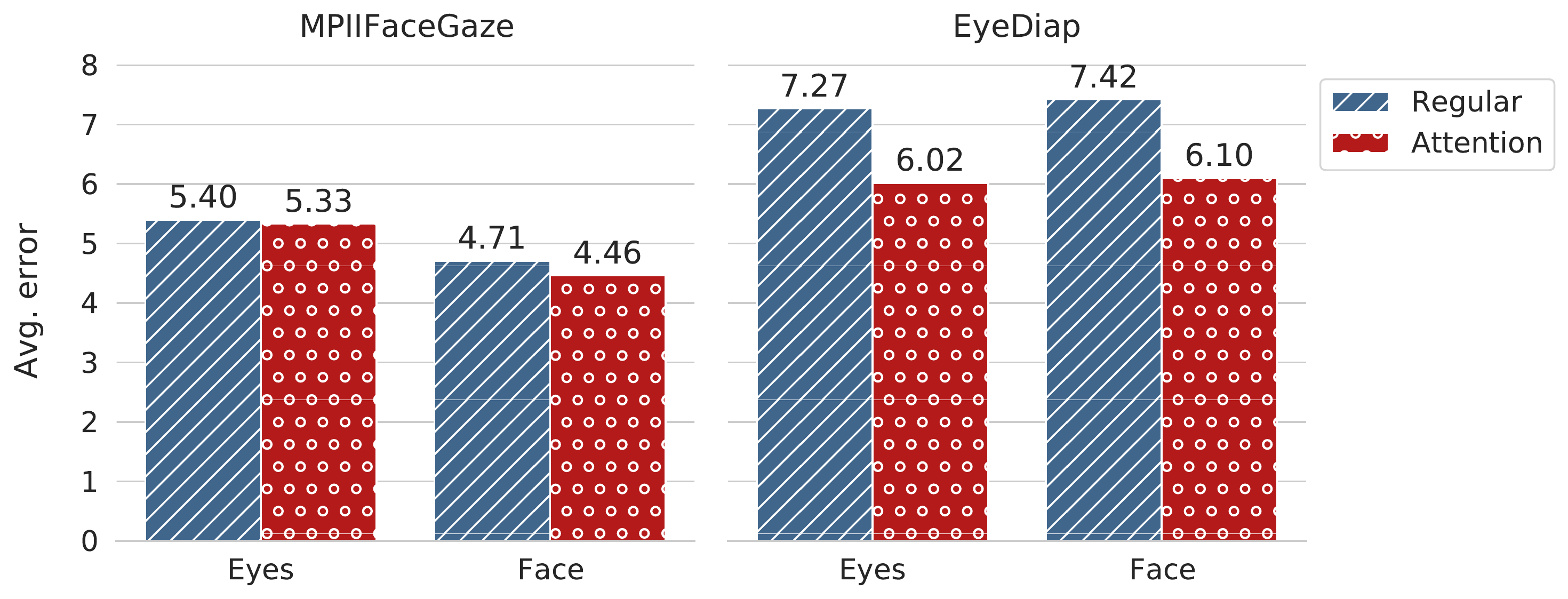}
\end{center}
\caption{Comparison of average angular error for single-branch gaze estimation networks on the MPIIFaceGaze and EyeDiap data sets. Blue bars represent evaluations with ResNet-14 as the backbone, while red bars represent those with ARes-14.}
\label{fig:eye_vs_face_plot}
\end{figure}

Figure \ref{fig:eye_vs_face_plot} shows a visual summary of the results of the single-branch networks from Table \ref{table:attablation}. There is a clear and consistent decrease of the average angular error in all instances when using the networks with self-attention augmented convolutions (ARes-14). As to which kind of input benefits the most from attention, on the EyeDiap data set, an error decrease of 17.19\% with eyes as input \textit{versus} 17.78\% with faces can be observed. On the MPIIFaceGaze data set, the decreases were of 1.28\% and 5.31\%, respectively. The larger magnitude of gains on the EyeDiap data set can be inferred from the fact that it is a more challenging data set with regards to head pose, which is an issue we guess self-attention augmentation benefits when applied to gaze estimation.

The slightly larger gains on the face-only network may be explained by the fact that while in eye images the region of interest is essentially the pupil, the regions of interest for gaze estimation in a facial picture (eyes, nose, mouth) are further apart from each other, which is exactly where self-augmentation can be the most useful. Although the error decrease in single-branch networks seems promising, results in Table \ref{table:attablation} shows that the advantage of using multiple-branches networks is very clear.

\begin{figure}[]
\begin{center}
\includegraphics[width=\textwidth]{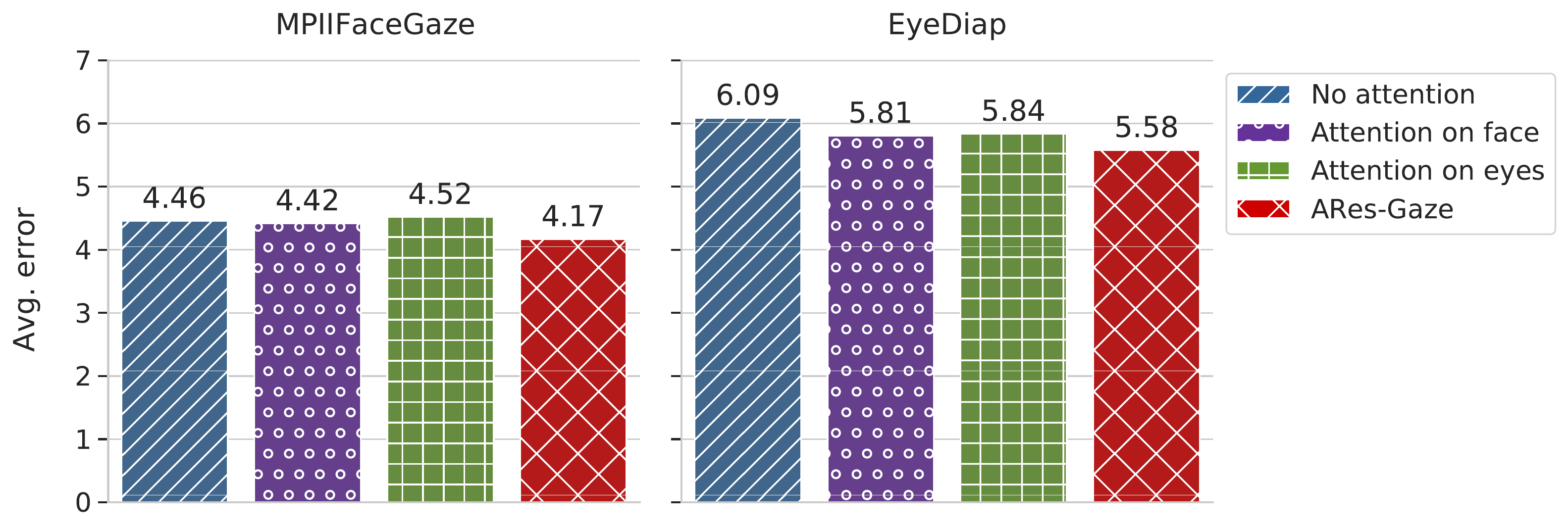}
\end{center}
\caption{Comparison of average angular error for four different versions of the proposed gaze estimation framework. From left to right, respectively: regular convolutional baseline (blue), a version with ARes-14 on the face branch and ResNet-14 on the eye branch (purple), a version with ResNet-14 on the face branch and ARes-14 on the eye branch (green), and the fully attention-augmented ARes-gaze (red).}
\label{fig:multi_input_plot}
\end{figure}

Figure \ref{fig:multi_input_plot} shows a comparison of the results obtained from multiple inputs across different iterations of our proposed gaze estimation architecture (replacing ResNet backbones by ARes-14 in each branch). It is worth noting that between the networks using ARes-14 as the backbone for only one of the branches, the one with self-attention augmentation on the face branch wins by a slight margin. On the MPIIFaceGaze data set, the one with attention only on the eye branch even had a small but noticeable drop in performance when compared with the regular CNN baseline. When analyzing the results from the single-branch network evaluation, it is possible to note that the face branch benefits slightly more from self-attention augmented convolutions due to having more distant elements that can be correlated by self-attention. This is reinforced by our results on the evaluation of mixed attention and regular convolution networks.

All in all, our findings indicate that self-attention augmented convolutions can be used as drop-in replacements to convolutional layers in gaze estimation networks to reduce the angular error in evaluation. Yet, while self-attention augmented convolutions work well with both face and eye-input images, our experiments showed that networks working with the full-face image as input were more prone to improvement when augmented by self-attention. The ARes-gaze framework which uses ARes-14 networks for both face and eye-inputs had the best results on our ablation studies, in some scenarios outperforming and at worst being comparable to state-of-the-art similar appearance-based gaze estimation methods on the MPIIFaceGaze and EyeDiap data sets.

\subsubsection{On the number of attention heads per convolutional layer}

\begin{figure}[]
\begin{center}
\includegraphics[width=\textwidth]{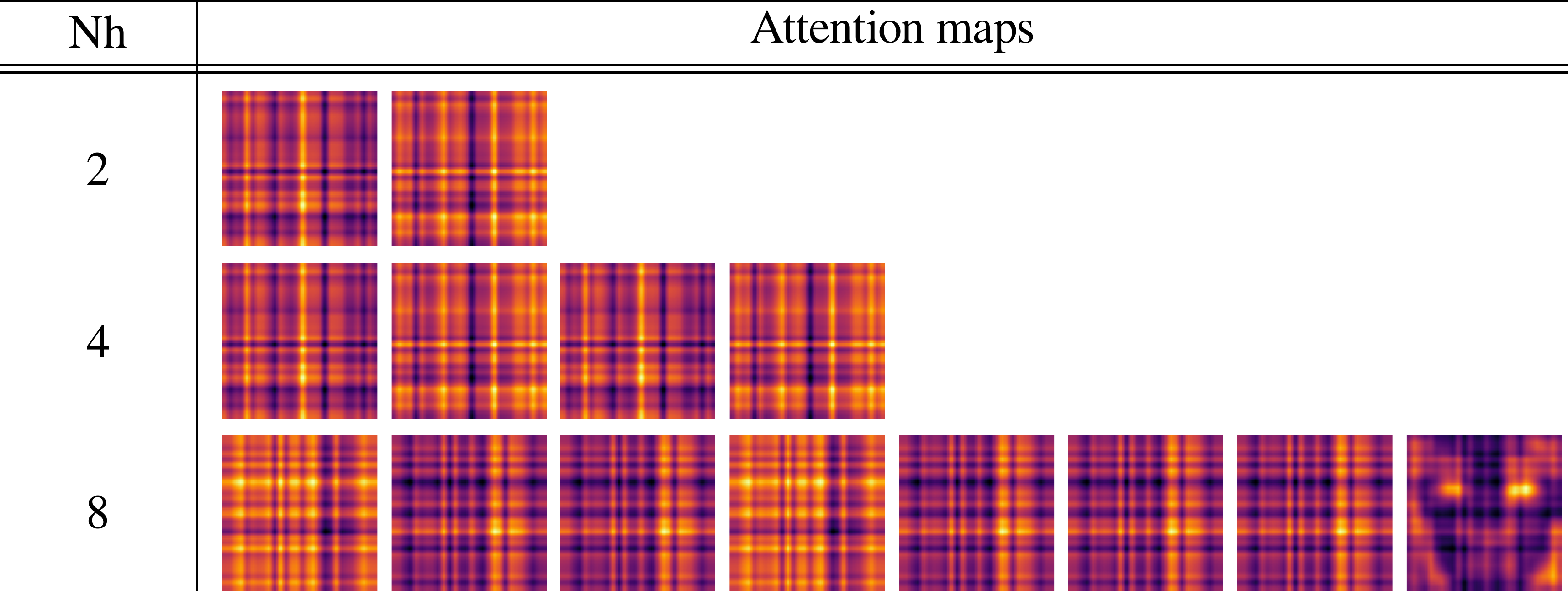}
\end{center}
\caption{Visualization of weights for intermediate feature maps from attention heads on the first attention-augmented convolution when performing inference.}
\label{fig:2-4-8attheads}
\end{figure}

\begin{figure}[t]
    \begin{subfigure}[]{0.2\textwidth}
        \includegraphics[width=\textwidth]{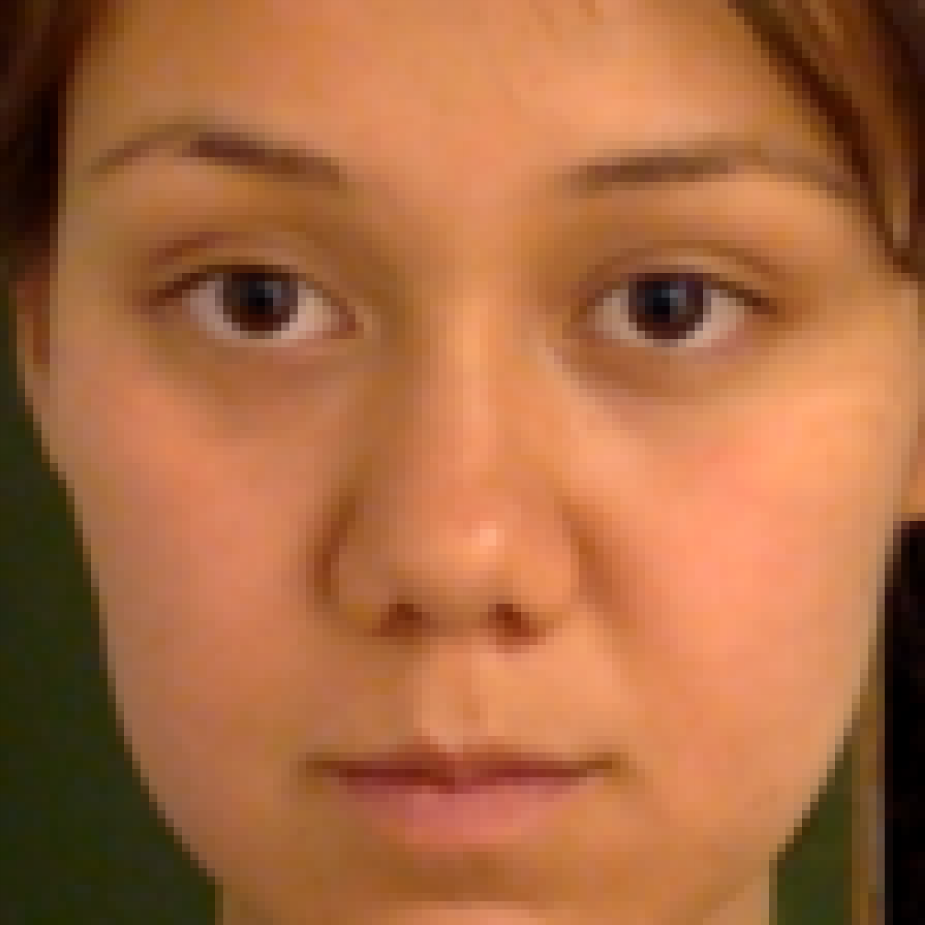}
        \caption{}
        \label{fig:sub:atthead_ablation_results_input}
    \end{subfigure}
    \begin{subfigure}[]{0.2\textwidth}
        \includegraphics[width=\textwidth]{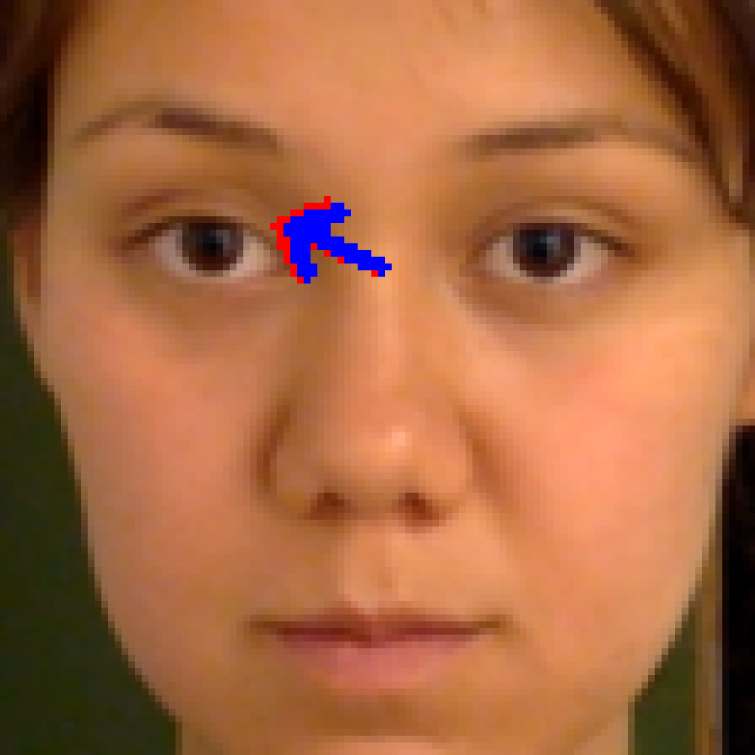}
        \caption{}
        \label{fig:sub:atthead_ablation_results_arrow}
    \end{subfigure}\\
    
    \begin{subfigure}[]{0.2\textwidth}
        \includegraphics[width=\textwidth]{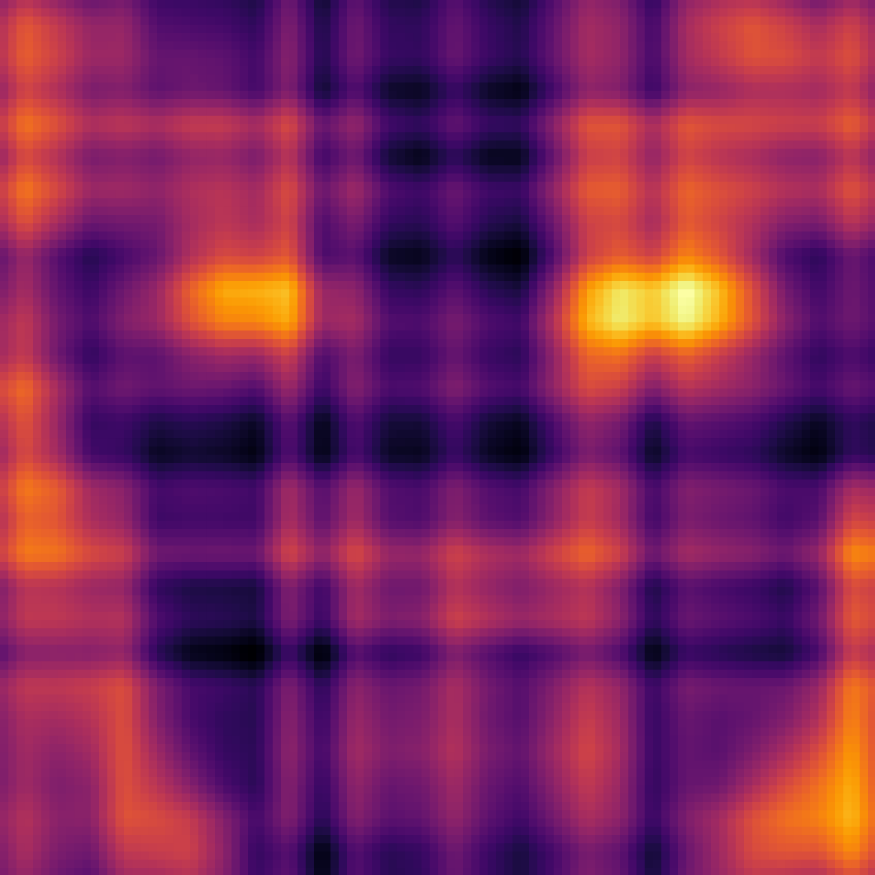}
        \caption{}
        \label{fig:sub:atthead_ablation_results_map}
    \end{subfigure}
    \begin{subfigure}[]{0.2\textwidth}
        \includegraphics[width=\textwidth]{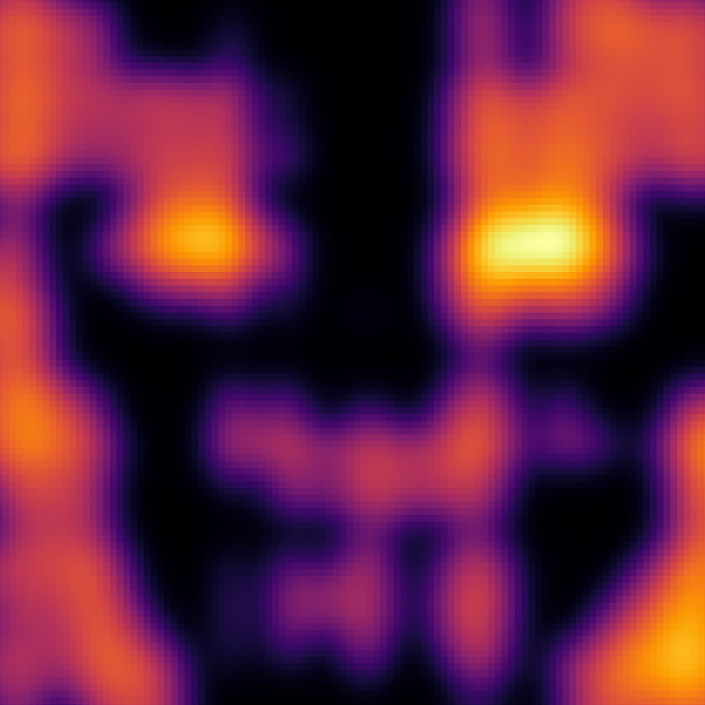}
        \caption{}
        \label{fig:sub:atthead_ablation_results_smoothmap}
    \end{subfigure}
    \begin{subfigure}[]{0.2\textwidth}
        \includegraphics[width=\textwidth]{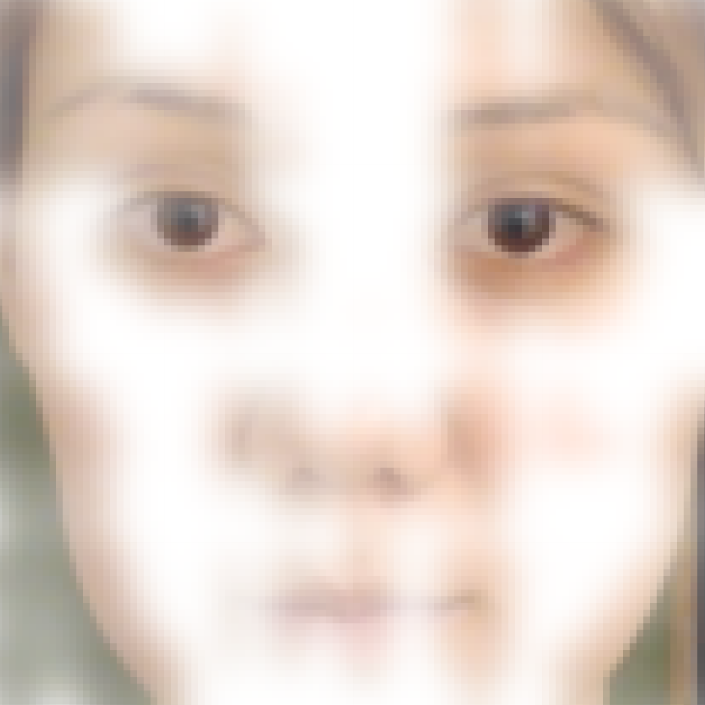}
        \caption{}
        \label{fig:sub:atthead_ablation_results_overlay}
    \end{subfigure}
\caption{Stages of attention maps: (a) is the input image, (b) is the projected prediction (red) and ground truth (blue) vectors, (c) is the last attention map on a random pixel in the image for the first convolutional layer, (d) is the attention map after thresholding and smoothing, and (e) is an overlay of the smooth map and the input image, evidencing relevant features.}
\label{fig:atthead_ablation_results}
\end{figure}

We obtained the counter-intuitive results that for a number of attention-heads less than eight, $Nh < 8$, the ARes-gaze framework sometimes actually performed worse than the regular convolutional baseline. As detailed in Section \ref{sec:attaugconv}, the output of a self-attention augmented convolution is the concatenation of a regular convolution and a self-attention layer, which is made up of feature maps obtained from the attention heads. We hypothesize that for lower numbers of attention-heads, the attention portion of the output may not be able to sufficiently represent relevant features, and may even harm the convolutional feature maps upon joining with them. This hypothesis is further reinforced by a visual analysis of early feature maps extracted from a trained network, as depicted in Fig. \ref{fig:2-4-8attheads}. Figure \ref{fig:2-4-8attheads} shows the weights computed from attention maps (one per attention head), extracted from the first augmented layer of an ARes-14 face branch during prediction. These weights are subsequently used to compute the final attention feature maps that will be joined with the convolution results for the output layer. Each row represents a different trained network with the number of attention-heads $Nh = 2, 4$ and $8$, as seen in Table \ref{table:attheadablation}. It is worth observing that, for some inputs, the self-attention augmented layer is capable of highlighting semantically relevant regions of the image. We verified however that when this happens, it is only on the map of the eighth attention head. This leads us to conclude that the attention layer might need a certain depth of attention-heads in order to specialize in very particular tasks. In turn, when this specialization is not reached properly, the overall output of the self-attention augmented convolution (concatenation of a regular convolution and an attention layer) can have lower quality than a regular convolutional layer with a larger feature space. This would explain the counter-intuitive results reached when using self-attention augmented layers with values lower than 8 for $Nh$, being less accurate than a regular convolutional CNN. 

To more easily interpret the highlighted regions by the weights of the eighth attention map, Fig. \ref{fig:atthead_ablation_results} illustrates an example of this phenomenon on a sample image from the MPIIFaceGaze data set. Figure \ref{fig:sub:atthead_ablation_results_arrow} shows the projected prediction and ground-truth gaze direction arrows over the input image (Fig. \ref{fig:sub:atthead_ablation_results_input}). Figures \ref{fig:sub:atthead_ablation_results_map}, \ref{fig:sub:atthead_ablation_results_smoothmap}, and \ref{fig:sub:atthead_ablation_results_overlay} show respectively the last attention map's weights for a random pixel on the input image, a refined version of these weights after operations of thresholding and smoothing, and an overlay with the input image. As expected, the eyes and eyebrows are the most relevant regions for the gaze estimation task. Notably, the nose, mouth, and background are also highlighted, which we speculate to be the source for head-pose related information implicitly used for the prediction.

\section{Conclusion}\label{sec:discussion}

In this paper, we addressed the question "can self-attention augmented convolutions be used to reduce angular error in appearance-based gaze estimation?", and we found that when compared to an equivalent regular convolutional network, the use of our 2D self-attention-based architecture can indeed produce more accurate results. We used ARes-14 twin branches as self-attention augmented CNNs in our experiments, and we believe that further research is merited on the design of optimal architectures for each branch of a multi-input attention-augmented framework such as the proposed ARes-gaze. We showed that the input face images had more to gain from using AAConvs than the input eye images, so incorporating domain knowledge of both attention mechanisms and gaze estimation to refine each branch for its particular input (face and eyes) may produce even better results than those reported. The spacial awareness afforded to the framework by the self-attention augmented convolutions can also be a promising way to develop joint head pose and gaze direction estimation networks, with the possibility of including other types of input images such as facial landmarks and explore their behavior in attention-augmented convolutional networks.

\section*{Acknowledgments}
The authors acknowledge the National Laboratory for Scientific Computing (LNCC/MCTI, Brazil) for providing HPC resources of the SDumont supercomputer, which have contributed to the research results reported within this paper. URL: http://sdumont.lncc.br. 

\section*{Funding}
This work was supported by the Fundação de Amparo à Pesquisa do Estado da Bahia (FAPESB) [grant number 892/2019]; and the Conselho Nacional de Desenvolvimento Científico e Tecnológico [grant number 307550/2018-4].

\bibliography{mybibfile}

\end{document}